\documentclass[accepted, specialissue]{melba}

\usepackage{xurl} 
\usepackage{gensymb}

\usepackage{amsmath,amsfonts}

\usepackage{comment}

\melbaid{2025:020}  
\doi{10.59275/j.melba.2025-8888}
\melbaauthors{Boland, Christopher and Tsaftaris, Sotirios and Dahdouh, Sonia}  
\email{christopher.boland@mre.medical.canon}
\volume{3}
\firstpageno{447}  
\melbayear{2025}  
\datesubmitted{2025/03}  
\datepublished{2025/11}  

\melbaspecialissue{Fairness of AI in Medical Imaging (FAIMI) 2025}
\melbaspecialissueeditors{Veronika Cheplygina, Aasa Feragen, Andrew King, Ben Glocker, Enzo Ferrante, Eike Petersen, Esther Puyol-Antón, Melanie Ganz-Benjaminsen}

\ShortHeadings{Preventing Shortcut Learning through Intermediate Layer Knowledge Distillation}{Boland, Tsaftaris, and Dahdouh}

\title{Preventing Shortcut Learning in Medical Image Analysis through Intermediate Layer Knowledge Distillation from Specialist Teachers}

\author{
	\firstname Christopher \surname Boland\aff{1,2},
	\name Sotirios A. Tsaftaris\aff{2},
	\name Sonia Dahdouh\aff{1},
    }
\affiliations{
	\num 1 \addr Canon Medical Research Europe, Edinburgh, EH6 5NP, UK \\
	\num 2 \addr School of Engineering, The University of Edinburgh, Edinburgh, EH9 3FG, UK \\
}

\abstract{
Deep learning models are prone to learning shortcut solutions to problems using spuriously correlated yet irrelevant features of their training data. 
In high-risk applications such as medical image analysis, this phenomenon may prevent models from using clinically meaningful features when making predictions, potentially leading to poor robustness and harm to patients.
We demonstrate that different types of shortcuts—those that are diffuse and spread throughout the image, as well as those that are localized to specific areas—manifest distinctly across network layers and can, therefore, be more effectively targeted through mitigation strategies that target the intermediate layers. 
We propose a novel knowledge distillation framework that leverages a teacher network fine-tuned on a small subset of task-relevant data to mitigate shortcut learning in a student network trained on a large dataset corrupted with a bias feature. 
Through extensive experiments on CheXpert, ISIC 2017, and SimBA datasets using various architectures (ResNet-18, AlexNet, DenseNet-121, and 3D CNNs), we demonstrate consistent improvements over traditional Empirical Risk Minimization, augmentation-based bias-mitigation, and group-based bias-mitigation approaches.
In many cases, we achieve comparable performance with a baseline model trained on bias-free data, even on out-of-distribution test data. 
Our results demonstrate the practical applicability of our approach to real-world medical imaging scenarios where bias annotations are limited and shortcut features are difficult to identify a priori.}


\begin{document}

\twocolumn[\maketitle]

\section{Introduction}
\enluminure{N}{eural} networks frequently demonstrate a preference for the path of least resistance during training, a phenomenon termed 
``simplicity bias'' \citep{NEURIPS2020_6cfe0e61}.
This tendency can lead these models to rely on features that, while strongly correlated with class labels in their training datasets, are irrelevant to the task. 
Such features, often referred to as ``shortcuts" or ``spurious correlations", yield an effective decision rule-set within the distribution of the training dataset but one which fails to generalize to data beyond this distribution \citep{geirhos2020shortcut}. 
For example, a model trained to identify cows in images may learn to detect grassy backgrounds rather than learning to understand what a cow looks like, if most training images show cows in grassy pastures. 
When presented with images of cows in novel contexts, such as on a beach, the model's prediction accuracy declines significantly \citep{beery2018recognition}. 
Because the decision rules learned by these systems prioritize such spurious features over robust, task-relevant ones, they fail to generalize to data where the spurious features are not available.
In contrast, a system that is trained to leverage reliable and task-relevant visual features should exhibit consistent performance even amidst shifts in data distribution.

In high-risk applications such as disease diagnosis, where clinical decisions rely on the accurate identification of subtle and often hard-to-detect disease features, shortcut learning represents a risk to patient safety.
Consider pneumothorax detection: popular chest X-ray datasets often contain images acquired post-treatment, once patients have been fitted with treatment devices like chest drains, which are visible in X-ray images.
Naturally, this creates a correlation between the presence of the treatment device and the disease label, which a network can learn, incorrectly, to use as a predictive feature of disease presence.
Consequently, the model is less accurate at detecting disease in patients who have not yet been treated \citep{murali2022shortcut}. Similarly, models trained to detect atelectasis in chest X-rays can learn to leverage the presence of ECG cables as a predictive feature \citep{olesen2024slicing}.
Disease detection models often rely inappropriately on such confounding features in addition to more subtle features including image acquisition protocol or even demographic characteristics \citep{souza2024identifying, konz2024reverse, seyyed2021underdiagnosis}. 
Sources of shortcut features in medical data are numerous, and their interactions are complex - exacerbating the challenge of monitoring and accounting for bias. 
This is magnified by the inconsistencies in metadata collection and labeling practices across datasets and healthcare institutions, making it impractical to track and account for all potential spurious features of the data. 

Emerging regulatory frameworks underscore the importance of these challenges. 
The European Union AI Act, due to come into effect in 2026, establishes comprehensive requirements for AI systems in high-risk domains like healthcare. 
The act mandates rigorous testing and monitoring of AI systems to identify and mitigate potential biases. 
Similarly, the World Health Organization's guidelines for AI in healthcare emphasize the need to safeguard patient safety and guarantee equitable treatment outcomes.
The FDA’s guidelines for AI and machine learning systems in healthcare applications necessitate detailed information regarding the metrics employed and how they ensure patient safety. 
Additionally, the guidelines request clarity on how to address any new or previously unidentified sources of bias that may arise, as well as details on how to disclose potential biases that could impact the model's effectiveness to users \citep{us2023marketing}. 
FDA approval for many AI systems in healthcare often requires a demonstration of ``practical equivalence'' to existing systems performing the same task, including evidence that the system's safety is on par with that of current processes \citep{petrick2023regulatory}.
These frameworks highlight the risks of deploying systems that may perpetuate or amplify existing healthcare disparities through learned biases. 
This regulatory landscape creates an urgent need for systematic approaches to identify and mitigate shortcut learning in medical AI systems. 

Current approaches to shortcut mitigation can be categorized according to their intervention point in the model development pipeline. 
Data-centric techniques address bias during pre-processing, where training data distributions are modified through resampling, reweighting, or augmentation to reduce imbalances with respect to bias features \citep{wu2023discover, li2019repair, ahmed2022achieving, pmlr-v174-zhang22a, liu2021just, wang2024drop, yun2019cutmix}. 
Model-centric techniques include (1) in-processing methods, which incorporate additional loss terms or penalties during training to discourage reliance on spurious features \citep{sagawa2019distributionally, muller2023localized, pmlr-v174-zhang22a, boland2024all} and (2) post-processing approaches, which attempt to remove learned biases from already trained models through fine-tuning or pruning \citep{xue2024bmft, wu2022fairprune, ghadiri2024xtranprune, bayasi2024biaspruner}.
A critical limitation across many of these methods is their dependency on accurate bias annotations for all training data. 
The assumption of access to comprehensive and reliable bias labels presents significant practical challenges in medical contexts, where the sources of bias are often numerous, interrelated, and difficult to identify a priori. 
Even when bias sources are known, obtaining accurate labels across diverse healthcare institutions with inconsistent metadata collection practices is prohibitively resource-intensive, limiting the real-world applicability of these approaches \citep{banerjee2023shortcuts}. 
Consequently, there is a need for methods that can address or reduce this burden of bias annotation while maintaining mitigation efficacy.

Recently, knowledge distillation (KD) has shown potential as a promising in-processing approach for preventing bias learning \citep{boland2024all,cha2022domain,bassi2024explanation,kenfack2024adaptive}. KD was originally proposed as a model compression technique where a smaller student network learns to mimic the predictions of a larger teacher network through an additional loss term that minimizes the divergence between the student model's outputs and those of the teacher model \citep{hinton2015distilling}. 
In the context of shortcut learning, a teacher trained on carefully curated data might help guide a student away from spurious correlations present in larger, potentially biased datasets.
Traditional knowledge distillation approaches typically focus only on matching the final layer outputs.
However, several works have demonstrated that learned biases can be detected in the intermediate layers of neural networks and can even be localized to specific network layers \citep{boland2024there, glocker2023risk, stanley2025and}.
Distillation approaches for debiasing that target the intermediate layers of the network may be able to mitigate biases more effectively.

In our previous work \citep{boland2024all}, we introduced an oracle-guided training approach to mitigate shortcut learning using a ``specialized teacher'', a model trained specifically on task-relevant, bias-free data.  While demonstrating promising results, this approach relied on matching batch-wise class probability distributions for knowledge transfer—a technique sensitive to batch composition. Here, we significantly extend this foundation through several methodological improvements. We replace batch-wise probability matching with sample-level Kullback-Leibler (KL) divergence between teacher and student predictions. This more principled approach provides direct guidance for each sample. We also extend the original framework to incorporate knowledge distillation in the final classification layer, complementing the intermediate layer guidance.
Furthermore, we significantly expand the experimental validation with extensive evaluation on out-of-distribution (OOD) test sets, systematic analysis of partial-layer distillation, utilization of compact teacher architectures to guide larger student networks, and evaluation of our method's efficacy when training data is corrupted with multiple, simultaneous shortcuts. 
All of these extensions serve to demonstrate the enhanced generalizability and practical applicability of our approach. 
Experiments across several network architecture designs, such as AlexNet, ResNet-18, DenseNet-121, and a lightweight 3D CNN, and over multiple medical image analysis tasks in different modalities, demonstrate that our approach is not modality-, task-, or architecture-specific.  
To further strengthen the viability of the approach in real-world scenarios, we validate our method on a recently released synthetic brain MRI dataset featuring subtle structural bias features, which are hard to detect upon simple inspection \citep{stanley2023flexible}.

Our proposed approach utilizes a teacher model trained on a small, carefully curated dataset to guide a student network's learning on larger, potentially biased datasets. By distilling knowledge at intermediate network layers, we encourage the student to learn robust, task-relevant features rather than relying on spurious correlations. Our \textbf{contributions} are as follows:
\begin{enumerate}
\item We demonstrate that intermediate-layer knowledge distillation from a teacher fine-tuned on a small amount of unbiased, task-relevant data effectively mitigates shortcut learning of a student trained on bias-corrupted data and leads to improved generalization as demonstrated through validation on out-of-distribution (OOD) test data. 
\item We provide empirical evidence that distillation at intermediate network layers significantly improves bias mitigation compared to final-layer distillation alone. 
\item We show that fine-tuning the teacher on task-relevant data leads to performance gains and reductions in bias compared to alternative distillation approaches, such as using a teacher pre-trained on ImageNet data or through confidence regularization of the intermediate layers. 
\item We establish that compact teacher architectures (e.g., AlexNet) can effectively guide larger student networks with different architectures (e.g., ResNet-18), critical for real-world deployment where much larger models, which would overfit to small, bias-free training subsets, are likely to be used.
\end{enumerate}

\section{Related Works}
This section reviews relevant literature in two key areas related to our work: approaches that address shortcut learning in deep neural networks and knowledge distillation techniques that can be leveraged for bias mitigation. We first explore various shortcut mitigation strategies and then examine how knowledge distillation can be adapted to address this challenge.

\subsection{Shortcut mitigation}
Shortcut mitigation techniques are grouped according to whether they modify the training data or the model's learning process. We review both data-centric and model-centric approaches, highlighting their respective strengths and limitations.
\subsubsection{Data-centric techniques}
Data-centric approaches address bias at the source by modifying training data distributions. 
Common approaches include up-sampling and down-sampling the dataset to remove the imbalance in the data with respect to the bias features, or re-weighting the loss to reduce the influence of the bias \citep{wang2020towards, sagawa2019distributionally}. 
Such approaches require bias labels and sufficient data diversity after resampling or augmentation. 
In contrast, methods like Just Train Twice (JTT)  \citep{liu2021just} and Discover and Cure \citep{wu2023discover} assign pseudo-labels of the bias feature to identify potentially biased samples before up-sampling or reweighting, avoiding the need for explicit bias annotations. 
These approaches estimate bias through model accuracy patterns or feature space representations.

Beyond basic resampling and re-weighting approaches, advanced data augmentation techniques have emerged as powerful tools for disrupting potential shortcuts. 
Cutout \citep{zhong2020random} introduces random occlusions by masking image regions, forcing models to learn more distributed representations. 
Mixup \citep{zhang2017mixup} creates synthetic training examples by interpolating between image pairs and their labels, reducing overfitting to training artifacts. 
CutMix \citep{yun2019cutmix} combines these approaches by replacing removed regions with patches from other training images.

While effective for natural images, these augmentation strategies face limitations in medical contexts. 
Disease features in medical images are often subtle and localized, unlike the prominent objects in natural image datasets. 
Random augmentations risk occluding critical diagnostic features, and they fail to target specific shortcut features systematically.

\cite{ahmed2022achieving} propose the use of a comprehensive pre-processing pipeline for pneumonia detection in chest X-rays involving normalization, region-of-interest (ROI) cropping, rotations, etc. 
Through evaluation with both IID and OOD test data, they validate that the influence of biases in the training data is significantly reduced compared to a model trained without applying this pre-processing. 
While this is relatively straightforward to implement, such an approach requires domain-specific tuning, knowledge of possible shortcut sources, and task-specific domain knowledge to inform some of the augmentation strategies, such as ROI cropping.

\subsubsection{Model-centric techniques}
Model-centric techniques mitigate learned biases by adjusting the model's weights and learning process, rather than targeting the training data itself.  
These can be further broken down into in-process techniques, which are applied at training time, and post-process techniques, which are applied after training is complete. 

Adversarial training methods \citep{correa2024efficient, zhang2018mitigating} introduce competing objectives to discourage reliance on biased features. 
However, these often require explicit labels for the bias sources and the competing objectives can introduce instability. 
Feature disentanglement techniques \citep{muller2023localized} attempt to separate task-relevant from spurious features but also often require explicit bias labels and rely on the assumption that such features are entirely task irrelevant, which may not always hold in practice. 

Post-processing methods like pruning \citep{wu2022fairprune} and fine-tuning \citep{xue2024bmft} attempt to remove shortcuts after training. 
Such approaches are particularly useful when it is not possible to re-train the model, for example, when the full, original training data is not available. 

\subsection{Knowledge Distillation}
While not originally developed for bias mitigation, knowledge distillation-inspired approaches to bias mitigation have shown promise in recent years.
\subsubsection{Traditional Knowledge Distillation}
Knowledge distillation, originally proposed as a method for model compression \citep{hinton2015distilling}, has recently shown effectiveness in mitigating bias learning \citep{boland2024all,cha2022domain,bassi2024explanation,kenfack2024adaptive}. 
Adopting a student-teacher training regime, knowledge from a large, well-trained teacher network is ``distilled'' into a smaller student network. 
Typically, this process incorporates an additional loss term that quantifies the divergence in predicted class probabilities between the two models  \citep{buciluǎ2006model}. 

\subsubsection{Distillation for bias mitigation}
\cite{tian2024distilling} demonstrate that distillation from a teacher trained on a balanced subset of training data can effectively mitigate learned biases in a student network trained on the full, biased dataset. 
However, they assume access to labels that accurately portray the source of bias in all of the teacher's training data, and they focus exclusively on the alignment of features in the final network layer, which may allow for more effective bias mitigation.
\cite{chai2022fairness} propose training a teacher model to overfit on its training data, and using its softened logits as training labels for a student model. 
The authors demonstrate that this soft labeling approach effectively functions as an error-based re-weighting mechanism that can improve fairness metrics without explicit demographic data. 
However, such an approach may not effectively capture all bias sources in the training data.

While traditional knowledge distillation focuses on final layer outputs, recent work has explored distillation at the intermediate layers. \cite{cha2022domain} propose MIRO. 
Utilizing a large pre-trained network as an ``oracle'' network, the authors formulate domain generalization as maximizing mutual information between the ``oracle'' model's representations and a target model's. 
Similarly, \cite{bassi2024explanation} propose ``explanation distillation'' as a technique to prevent shortcut learning in deep neural networks. 
Their approach distills explanations from a teacher model pre-trained on a massive, diverse dataset, but not necessarily one with task-specific knowledge. 
This lack of knowledge pertaining to the specific task of the student in the teacher network limits its ability to guide the student network to robust, task-relevant features.

\cite{boland2024all} introduced an oracle-guided training approach for shortcut mitigation that does not require explicit bias labels for the full training dataset of the student, nor does it make assumptions about bias characteristics. 
Our work builds upon this foundation through methodological improvements for enhanced robustness, exploration of knowledge distillation applied only to subsets of the student network's intermediate layers, and the use of low-capacity teacher networks to guide high-capacity students.

\section{Methods}
\begin{figure*}[t]
    \centering
    \includegraphics[width=\textwidth]{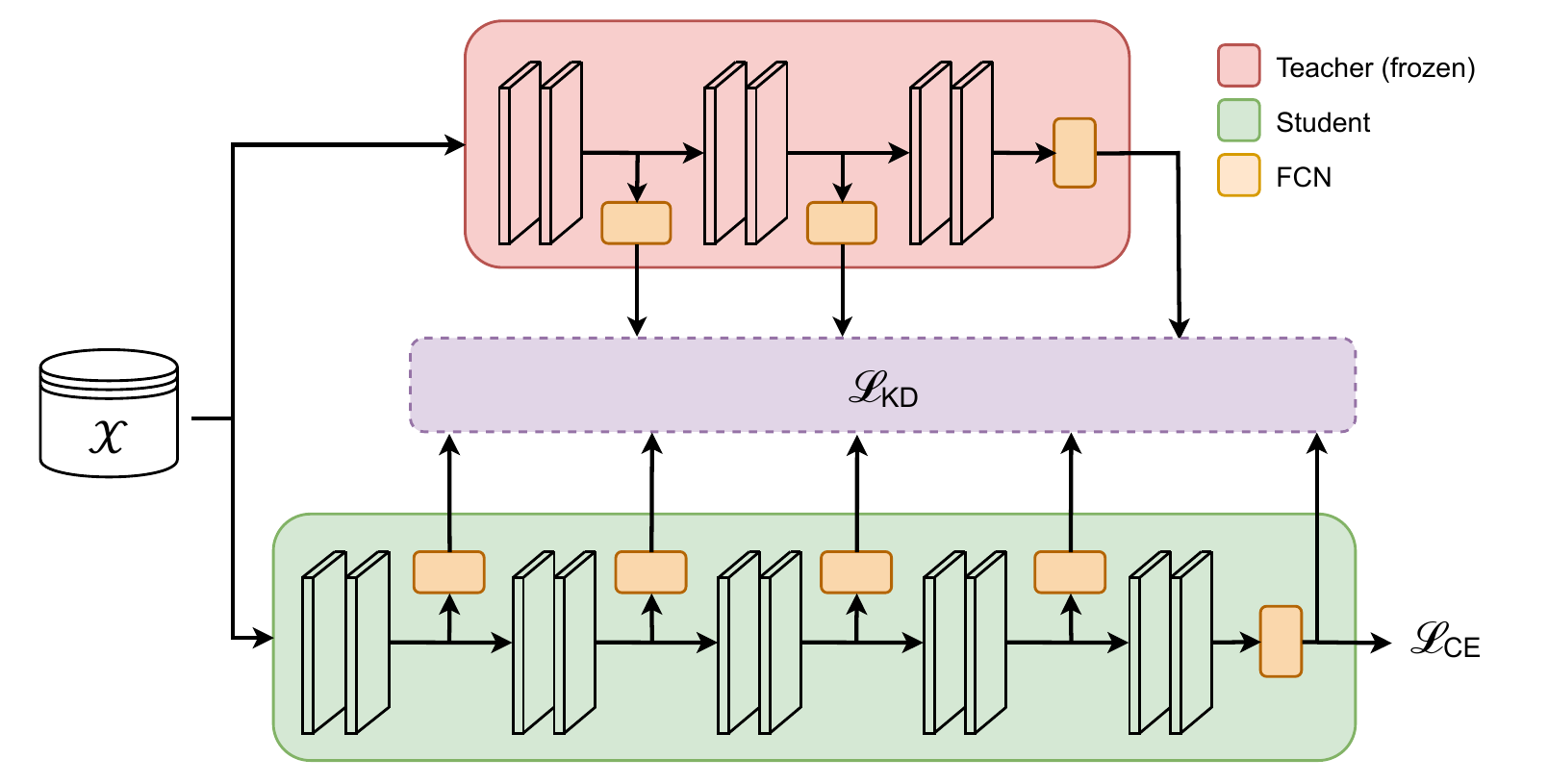}
    \caption{Overview of the proposed student-teacher training method. 
    The teacher network, trained on clean data, guides the student model's learning process through the distillation of task-specific knowledge to the intermediate layers.}\label{fig:architecture}
\end{figure*}

Our proposed approach addresses the challenge of shortcut learning through a novel teacher-student knowledge distillation framework that guides feature learning at multiple network depths (Figure \ref{fig:architecture}). 
Central to our approach is the observation that the influence of shortcut learning is detectable in a network's intermediate layers, suggesting that effective mitigation strategies should target the entire network rather than just the final output \citep{boland2024there}.
 
In this section, we present our method for measuring intermediate-layer model confidence. We then detail our knowledge distillation approach for mitigating shortcut learning, followed by our experimental setup, including datasets, synthetic shortcut designs, and evaluation protocols.

\subsection{Model confidence and shortcut learning}
Models trained on biased data tend to exhibit overconfidence in their predictions \citep{utama2020mind}. 
Shortcut features in a model's training data provide an easier decision rule-set with which to infer class. 
These simple features allow the model to achieve high confidence with less effort \citep{ao2023confidence}. 
Prior work has demonstrated that these spurious features lead to detectable changes in the internal behavior of the network \citep{boland2024there}.
We are interested in (a) the confidence with which a trained network infers class through the internal layers, (b) how training on shortcut-corrupted data changes this behavior, and (c) if knowledge distillation from an unbiased teacher can mitigate shortcut learning. 
To understand this, we introduce classification probes (linear classification heads, consisting of an average pooling layer and a single fully connected layer) which are attached to the intermediate layers of both the student and teacher networks. 
After the network finishes training, these probes are fine-tuned on the downstream task. 
Once trained, the probes offer insight into a model's ability to infer the true class at different depths of the network, in addition to facilitating knowledge distillation from the teacher network's intermediate layers to the student's. 

\subsection{Measuring Model Confidence}
To quantify a model's confidence over a batch at each layer, we follow prior work and consider the output logits of the classification probes \citep{taha2022confidence}. 
The sigmoid of the output logits ranges from  $0$ to $1$, indicating the likelihood that the input belongs to the positive ($1$) or negative ($0$) class. 
We quantify model confidence $C(\mathcal{X})$ as the deviation from maximum uncertainty ($0.5$), where higher values indicate greater prediction certainty.
This is illustrated in Equation \ref{eq:conf}, where $f(x)$ represents the sigmoid output of the model for input $x$.

\begin{equation}\label{eq:conf}
C(\mathcal{X}) = \sum_{x \in \mathcal{X}}|f(x) - 0.5|
\end{equation}

\subsection{Mitigation of shortcut learning via knowledge distillation}
Our training scheme (Fig.~\ref{fig:architecture}) aims to mitigate shortcut learning by preventing the student model from becoming overconfident through the use of shortcut features. 
The student model is trained to minimize the cross-entropy loss on a biased dataset while matching the teacher network's class probabilities at each layer.

\subsubsection{Teacher-Student Architecture}
The ``specialist teacher'' model is defined as a network trained on a small, carefully curated subset of the full training dataset. 
This subset is manually selected to contain balanced class representation and to be free of the bias features present in the student's training data. 
Unlike traditional knowledge distillation approaches that use large, general-purpose teachers, our specialized teacher possesses task-specific knowledge while avoiding the spurious correlations that contaminate larger datasets.

Importantly, all samples used to train the teacher model are excluded from the student's training dataset to prevent leakage between the teacher and the student's training data. 
For the teacher model, we follow a standard training procedure: the network is trained to completion, then frozen before the classification probes are fine-tuned on the downstream task. 
We use separate optimizers for updating the network parameters and the probe parameters to prevent unintended interactions between their learning objectives.

The student model is trained on the biased dataset using both the standard classification loss and the knowledge distillation from the teacher. 
At each epoch, after the student model's parameters have been updated, the network's encoder and final classification head are frozen, and the probes are fine-tuned on the downstream classification task. 
This maintains the probes' ability to classify based on the student's currently learned feature embeddings while preventing undesired interaction between the probe training and the student's feature learning.

We encourage alignment between teacher and student by minimizing the KL divergence between the output probability distributions of each model's intermediate layer classification probes. 
Following other intermediate-layer knowledge distillation literature \citep{haidar2021rail, bassi2024explanation}, we also apply knowledge distillation loss on the output of the network's final classification head.

\subsubsection{Loss Functions}
The training loss of the student is described in Eq. \ref{eq:encoder_loss}, where $\mathcal{L}_{total}$ is the total loss, $\mathcal{L}_{CE}$ is the Cross Entropy (CE) loss, $\mathcal{L}_{KD}$ is the knowledge distillation loss between the teacher and student probes, and $\lambda_{i}$ is a weight applied to each loss to allow the trade-off between each objective to be managed. 
For simplicity, we set all weights equal to 1. 
KL divergence loss is defined in Eq. \ref{eq:kl_loss}
 where we have two sets of intermediate layer predictions, $S = \{p_1, p_2, ..., p_n\}$ and $T = \{q_1, q_2, ..., q_n\}$ where $S$ represent the set of intermediate layer outputs of the student network, $T$ represents the teacher, and $\alpha_{i}$ represents the weight of the distillation loss to the $i^{th}$ layer of the student.
 
\begin{equation}\label{eq:kl_loss}
    \mathcal{L}_{KD} = \sum_i^n \alpha_{i}D_{KL}(p_i^S || q_i^T)
\end{equation}

\begin{equation}\label{eq:encoder_loss}
    \mathcal{L}_{total}=\lambda_{1} \cdot \mathcal{L}_{CE} + \lambda_{2} \cdot \mathcal{L}_{KD}
\end{equation}

\subsection{Experimental Setup}
\subsubsection{Datasets}
\begin{table}[!ht]
\centering
\caption{Composition of positive/negative class samples in train, validation, and test splits of our datasets.}
\begin{tabular}{c c c c}
\midrule
Dataset & Train & Valid & Test \\
 \midrule
 CheXpert & 1457/1457 & 365/406 & 600/300 \\
 ISIC & 560/560 & 140/146 & 207/393 \\
 SimBA & 1291/1292 & 323/323 & 530/544 \\
 MIMIC & n/a & n/a & 500/500 \\
 Fitzpatrick17k & n/a & n/a & 69/69\\
 \bottomrule
\end{tabular}

\label{tab:datasets}
\end{table}

We evaluate our proposed method using three medical imaging datasets of different modalities and tasks. For each, we enforce class balancing by downsampling the majority class and combine the original train/validation splits. New splits are generated when we run k-fold cross-validation. Table \ref{tab:datasets} summarizes the composition of positive/negative class samples across train, validation, and test splits for each dataset.
\begin{itemize}
\item CheXpert \citep{irvin2019chexpert}: 
a large-scale chest radiography dataset comprised of $224,316$ chest X-rays from $65,240$ patients and $14$ disease labels.
In our experiments, we consider the task of pneumothorax detection. 
We use a subset of the full CheXpert dataset containing an equal number of pneumothorax-positive and no finding-positive images. 
Our final training dataset consists of $2,914$ images.
\item ISIC 2017 \citep{codella2018skin}: a popular skin lesion image dataset from the International Skin Imaging Collaboration containing $2,000$ dermoscopic images. 
We perform binary classification between malignant lesions (melanoma/seborrheic keratosis) and benign lesions. 
Our training split contains $1,120$ images after class balancing.

\item SimBA \citep{stanley2023flexible}: a fully synthetic brain MRI dataset which allows evaluation on 3D medical data with controlled biases. 
It contains simulated structural changes associated with the class label alongside artificially introduced morphological deformations as potential shortcuts. 
We also utilize the original version of the dataset without any bias features added.

\end{itemize}

To assess the generalization of trained models, we include two out-of-distribution (OOD) datasets for evaluation:
\begin{itemize}
\item MIMIC \citep{goldberger2000physiobank, johnson2024mimic, johnson2019mimic}: for CheXpert evaluation, we use a class-balanced evaluation set composed of pneumothorax-positive and no finding samples from the MIMIC dataset, another large-scale chest radiograph dataset acquired from a different institution.

\item Fitzpatrick17k \citep{groh2021evaluating, groh2022towards}: for ISIC evaluation, we leverage a second dermatological image dataset.
\end{itemize}

\subsubsection{Synthetic shortcuts}
\begin{figure}[!th]
\centering
\includegraphics[width=0.5\textwidth]{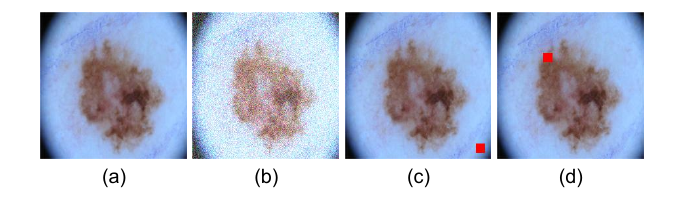}
\caption{ISIC skin lesion image augmented with synthetic shortcuts: (a) original; (b) noise; (c) square (constant location); (d) square (random location). The noise effect has been amplified here for illustrative purposes.
}\label{fig:shortcut examples}
\end{figure}

Inspired by research highlighting common shortcut sources in medical image analysis datasets such as acquisition devices, scanning protocols, hospital tags, and medical devices,
 
we design a controlled environment for the empirical evaluation of our approach. We introduce synthetic bias features into our ISIC and CheXpert training datasets that allow us to assess the generalizability of our method across diverse types of bias. We design several experimental setups featuring a single bias feature and multiple, concurrent bias features. We augment our datasets with one of three unique synthetic bias features (Figure ~\ref{fig:shortcut examples}): 
\begin{enumerate}
    \item \textbf{Diffuse:} leveraging random, uniform noise patterns as a spurious signal spread throughout the image. The noise is generated using a uniform distribution with values between 0 and 0.15 applied to each pixel. Such shortcut features are designed to simulate those that may be caused by acquisition devices and scanning protocols \citep{ong2024shortcut}.
    \item \textbf{Localized:} introducing small square shapes to the image, similar to other work \citep{dagaev2023too},  we aim to simulate more localized shortcut features of various complexities seen in the literature, such as hospital tags and treatment devices \citep{olesen2024slicing}.  We test two variants: 
    \begin{enumerate}
        \item \textbf{Constant location:} the square appears in a fixed spot.
        \item \textbf{Random location:} the location of the square varies among images.
    \end{enumerate}
\end{enumerate}

\begin{figure}[t]
    \centering
    \includegraphics[width=0.5\textwidth]{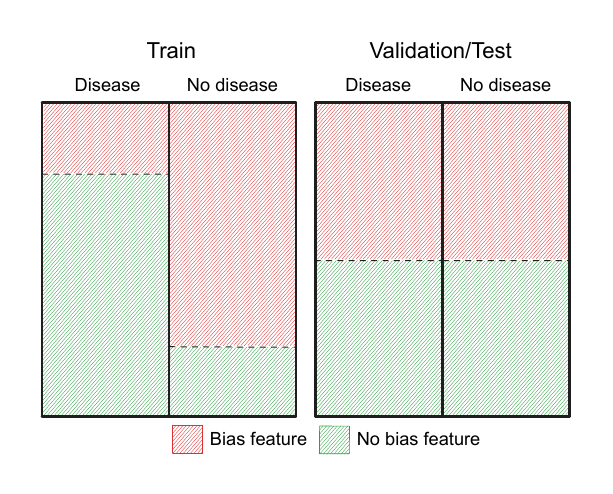}
    \caption{Illustrative representation of the synthetic shortcut feature distribution in our train, validation, and test splits in the CheXpert and ISIC datasets.}\label{fig:shortcut_balance}
\end{figure}

In our training splits, the shortcut features are correlated with the class label (Figure \ref{fig:shortcut_balance}).
We vary the prevalence of the shortcut feature (the degree of its correlation with the class label in the training data) to assess its influence on training and mitigation efforts. 
In the validation and test splits, shortcut features are balanced across both classes.
In cases with two simultaneous shortcut features, each is correlated with a different class label. 

Notably, in the case of the SimBA dataset, all data splits exhibit the same bias prevalence.
Experiments on SimBA allow us to validate the efficacy of our method when it is not possible to access an unbiased validation set.

\subsection{Evaluation metrics and statistical analysis}
We evaluate our experiments by considering both overall performance metrics and bias-specific metrics. 
For classification performance, we report the Area Under the Receiver Operating Characteristic Curve (AUC), providing a threshold-free measure of discriminative ability across all datasets. 
Similar to recent research investigating bias and shortcut learning in medical image analysis, we quantify the impact of shortcut features on model predictions through True Positive Rate disparity ($\Delta$TPR) \citep{glocker2023risk, stanley2025and}. 
$\Delta$TPR directly measures the model's ability to maintain consistent sensitivity across bias-aligned and bias-contrasting groups, a critical requirement for clinical deployment where missed diagnoses (false negatives) carry severe consequences. 
We used a 0.5 classification threshold in all cases as the natural decision boundary for binary classification.

We define bias-aligned samples as those whose combination of class label and shortcut feature presence matches the class-bias correlation established in the training data. 
Meanwhile, bias-contrasting samples represent those whose combination of class label and shortcut feature presence opposes the class-bias correlation in the training data. 
Statistical significance between performance differences is assessed using paired t-tests with Bonferroni correction to account for multiple comparisons where appropriate.

\subsection{Benchmark methods}
We compare our approach with several established shortcut learning mitigation methods. The network trained on the original, clean dataset without shortcut features is our \textbf{Baseline}. The network trained on the shortcut-corrupted dataset with standard cross-entropy optimization is referred to as \textbf{ERM}. We compare to four augmentation-based approaches: \textbf{CutOut} \citep{zhong2020random}, \textbf{MixUp} \citep{zhang2017mixup}, \textbf{CutMix} \citep{yun2019cutmix}, as well as the use of random rotation (up to $15\degree$) and horizontal flip augmentations \textbf{(Aug)}. We also compare to two popular group-based methods: \textbf{GroupDRO (GDRO)} \citep{sagawa2019distributionally} and \textbf{Just Train Twice (JTT)} \citep{liu2021just}

For each method, we implement configurations following the authors' recommendations without any specific fine-tuning or adjustments made for our data and use identical architecture backbones for fair comparison.

\subsection{Implementation Details}\label{sec:imp_details}
In all experiments, we utilize an AdamW optimizer with weight decay of $0.1$ and train our models with a learning rate of $1{\times}e^{-4}$. All intermediate layer classification probes are trained with a learning rate of $0.1$. 
For our 2D datasets (ISIC, CheXpert, MIMIC, and Fitzpatrick17k) images are re-sized to ImageNet resolution, $224\times224$, while for SimBA, we resize to $96\times96\times96$. 
We do not apply any rotation or flipping augmentations by default. 
We set the maximum number of training epochs to 1000 with early stopping after 15 epochs if there is no improvement in the validation loss. 
In none of our experiments does training run for the complete 1000 epochs without reaching the early stop condition.
We use 5-fold cross-validation, with consistent test sets across folds. 
Our experimental setup utilizes Python and PyTorch, and we train on an NVIDIA RTX 2080 Ti and Tesla V100s.

For the 3D experiments on SimBA data, we employ a lightweight 3D CNN consisting of five convolutional blocks, each containing a 3D convolutional layer (kernel size 3×3×3), batch normalization, and Sigmoid activation. Classification probes are attached after each convolutional block, consisting of 3D global average pooling followed by a linear layer.

\begin{figure*}[!th]
    \centering
    \includegraphics[width=\textwidth]{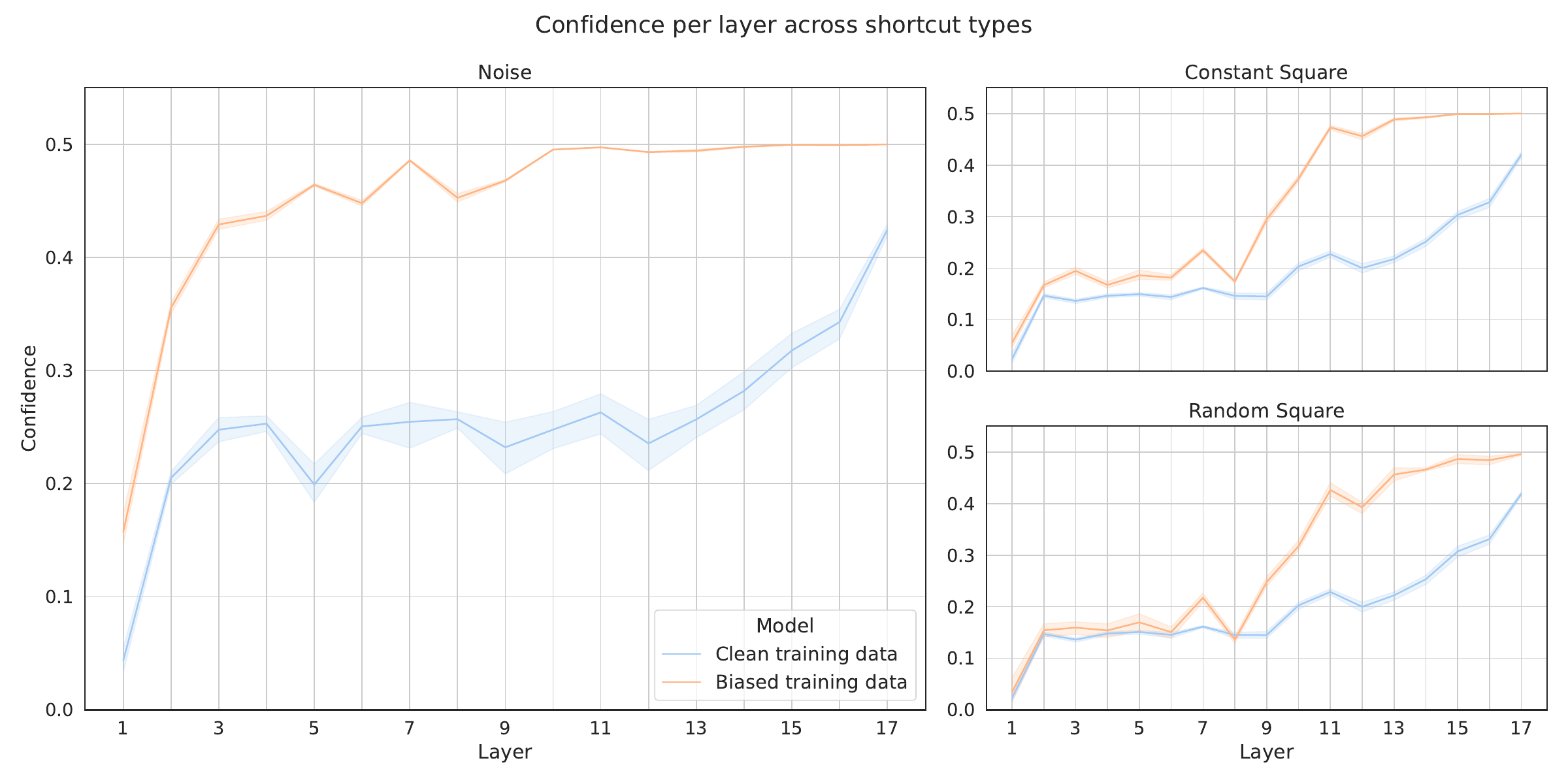}
    \caption{Intermediate-layer confidence of two ResNet-18 models trained on near-identical training sets. Confidence bands represent the standard deviation over 5-fold cross-validation. Both networks are trained on the CheXpert dataset with a learning rate of $1e^{-4}$. Intermediate layer classification probes have a learning rate of $0.1$. The training data of one model has been corrupted with various synthetic shortcut features, while the training data of the other has not.}\label{fig:per_layer_conf_of_shortcuts}
\end{figure*}

\section{Results}
Our experimental evaluation examines several key aspects of the proposed approach. We first investigate how shortcut learning manifests in intermediate network layers, then evaluate our knowledge distillation method against alternative approaches. We also analyze the impact of partial layer distillation, the effectiveness of compact teacher architectures, and performance on realistic structural biases in 3D medical data. In the following work, our teacher networks are trained on a subset of 20\% of the full, original training data. The samples in this subset are removed from the student network's training data. We later explore the efficacy of our teacher with fewer training data.

\subsection{Core method validation}

We begin our experimental evaluation by establishing fundamental evidence for our approach: first demonstrating how shortcut learning manifests in neural networks, then validating our knowledge distillation method's effectiveness across diverse experimental conditions.

\subsubsection{Shortcut learning manifests distinctly across network layers }\label{sec:shortcut_ilc}
Considering the complexity of many medical image analysis tasks, we expect a well-trained model using clinically relevant features to exhibit lower confidence than a model relying on easy shortcut features. Additionally, we might expect that the confidence of a model reliant on shortcut features will increase in earlier layers, aligning with the expectation that the deeper layers of the network capture more sophisticated features \citep{baldock2021deep}. 

To test this, we train two ResNet-18 models on CheXpert following Empirical Risk Minimization (ERM), where we simply aim to optimize cross-entropy loss. One is trained on the original dataset without any synthetic biases, while the other is trained on the same dataset augmented with synthetic shortcut features associated with the disease class. In this case, the shortcut features have a 100\% prevalence rate (perfectly correlated with the disease class). After training, we fine-tune our classification probes for each model. Each model is evaluated on our held-out test set, and the output of the probes are used to evaluate the influence of the bias feature on the network's predictive behavior.

Figure ~\ref{fig:per_layer_conf_of_shortcuts} illustrates the per-layer confidence of each model. 
In line with our hypothesis, the model trained on the biased data becomes overconfident compared to the baseline trained on clean data. 
In the case of our diffuse noise shortcut, we observe this to an extreme degree in the earliest layers, while the localized shortcuts don't result in a large degree of overconfidence until the later layers. 
This is likely because the diffuse shortcut is composed of low-level features that require very little disambiguation by the network, unlike the localized shortcuts. 
We observe similar patterns in other datasets and network architectures that we evaluate.

\subsubsection{Intermediate-layer knowledge distillation mitigates shortcut reliance}\label{sec:kd_main_res}

\begin{table*}[!ht]
\centering
\caption{$\Delta$TPR $\downarrow$ between bias-aligned and bias-contrasting samples for a ResNet-18 trained on data with various bias prevalence rates. Results are presented as Mean$\pm$Std over 5-fold cross-validation. Models are marked as \textbf{best} and \underline{second-best}. When the difference between first and second best is statistically significant ($p<0.05$ according to a paired t-test), the best-performing model is highlighted  with *.}
\footnotesize
\begin{tabular}{c c c c c c c c}
\midrule
Prev. & \multirow{2}{*}{Model} & \multicolumn{3}{c}{CheXpert} &\multicolumn{3}{c}{ISIC} \\ \cmidrule(lr){3-5} \cmidrule{6-8}
 (\%) & & Noise & Square (C) & Square (R) & Noise & Square (C) & Square (R) \\
 \midrule
 0 & Baseline & 0.131\scriptsize$\pm$0.051 & 0.020\scriptsize$\pm$0.008 & 0.015\scriptsize$\pm$0.013 & 0.409\scriptsize$\pm$0.095 & 0.056\scriptsize$\pm$0.015 & 0.050\scriptsize$\pm$0.017 \\
 
\midrule

\multirow{6}{*}{100}
 & ERM  & 1.000\scriptsize$\pm$0.000 & 1.000\scriptsize$\pm$0.000 & 0.991\scriptsize$\pm$0.008 & 1.000\scriptsize$\pm$0.000 & 0.777\scriptsize$\pm$0.093 & 0.844\scriptsize$\pm$0.168 \\
 & MixUp & 0.987\scriptsize$\pm$0.026 & 0.998\scriptsize$\pm$0.003 & 0.854\scriptsize$\pm$0.119 & \underline{0.998\scriptsize$\pm$0.004} & 0.875\scriptsize$\pm$0.152 & 0.754\scriptsize$\pm$0.157 \\
 & CutOut  & 0.999\scriptsize$\pm$0.001 & 1.000\scriptsize$\pm$0.000 & 0.832\scriptsize$\pm$0.112 & 1.000\scriptsize$\pm$0.000 & 0.448\scriptsize$\pm$0.065 & 0.277\scriptsize$\pm$0.064 \\
 & CutMix  & 0.993\scriptsize$\pm$0.005 & \underline{0.503\scriptsize$\pm$0.073} & \underline{0.126\scriptsize$\pm$0.022} & 1.000\scriptsize$\pm$0.000 & 0.359\scriptsize$\pm$0.063 & \underline{0.116\scriptsize$\pm$0.053} \\
 & Aug & \underline{0.957\scriptsize$\pm$0.092} & 0.979\scriptsize$\pm$0.013 & 0.984\scriptsize$\pm$0.006 & 1.000\scriptsize$\pm$0.000 & \underline{0.161\scriptsize$\pm$0.052} & 0.515\scriptsize$\pm$0.133 \\
 & Ours  & \textbf{0.377\scriptsize$\pm$0.185}* & \textbf{0.079\scriptsize$\pm$0.017}* & \textbf{0.035\scriptsize$\pm$0.013}* & \textbf{0.068\scriptsize$\pm$0.055}* & \textbf{0.034\scriptsize$\pm$0.016}* & \textbf{0.074\scriptsize$\pm$0.042} \\
 
\midrule

\multirow{8}{*}{95}
 & ERM & 0.939\scriptsize$\pm$0.028 & 0.912\scriptsize$\pm$0.086 & 0.791\scriptsize$\pm$0.093 & 0.959\scriptsize$\pm$0.019 & 0.861\scriptsize$\pm$0.036 & 0.703\scriptsize$\pm$0.103 \\
 & MixUp  & 0.927\scriptsize$\pm$0.105 & 0.987\scriptsize$\pm$0.008 & 0.693\scriptsize$\pm$0.103 & 0.952\scriptsize$\pm$0.037 & 0.936\scriptsize$\pm$0.035& 0.757\scriptsize$\pm$0.066 \\
 & CutOut & 0.950\scriptsize$\pm$0.050 & 0.912\scriptsize$\pm$0.069 & 0.636\scriptsize$\pm$0.131 & 0.946\scriptsize$\pm$0.047 & 0.579\scriptsize$\pm$0.249 & 0.439\scriptsize$\pm$0.146 \\
 & CutMix & 0.975\scriptsize$\pm$0.030 & \underline{0.325\scriptsize$\pm$0.048} & \underline{0.142\scriptsize$\pm$0.063} & \underline{0.922\scriptsize$\pm$0.079} & 0.311\scriptsize$\pm$0.107 & \underline{0.134\scriptsize$\pm$0.040} \\
 & Aug & \underline{0.899\scriptsize$\pm$0.072} & 0.779\scriptsize$\pm$0.150 & 0.813\scriptsize$\pm$0.061 & 0.935\scriptsize$\pm$0.054 & \underline{0.205\scriptsize$\pm$0.079} & 0.549\scriptsize$\pm$0.115 \\
 & GDRO & 0.986\scriptsize$\pm$0.010 & 0.978\scriptsize$\pm$0.015 & 0.702\scriptsize$\pm$0.091 & 0.967\scriptsize$\pm$0.014 & 0.800\scriptsize$\pm$0.035 & 0.477\scriptsize$\pm$0.029 \\
 & JTT  & 0.982\scriptsize$\pm$0.011 & 0.946\scriptsize$\pm$0.031 & 0.673\scriptsize$\pm$0.065 & 0.946\scriptsize$\pm$0.031 & 0.793\scriptsize$\pm$0.043 & 0.505\scriptsize$\pm$0.073 \\
 & Ours & \textbf{0.372\scriptsize$\pm$0.110}* & \textbf{0.089\scriptsize$\pm$0.023}* & \textbf{0.047\scriptsize$\pm$0.027} & \textbf{0.077\scriptsize$\pm$0.069}* & \textbf{0.100\scriptsize$\pm$0.039} & \textbf{0.052\scriptsize$\pm$0.012}* \\

\midrule

\multirow{8}{*}{85}
 & ERM & 0.745\scriptsize$\pm$0.145 & 0.735\scriptsize$\pm$0.113 & 0.423\scriptsize$\pm$0.063 & \underline{0.274\scriptsize$\pm$0.037} &  0.376\scriptsize$\pm$0.094 & 0.294\scriptsize$\pm$0.083 \\
 & MixUp & 0.699\scriptsize$\pm$0.115 & 0.677\scriptsize$\pm$0.131 & 0.336\scriptsize$\pm$0.028 & 0.490\scriptsize$\pm$0.091 & 0.523\scriptsize$\pm$0.100 & 0.407\scriptsize$\pm$0.075 \\
 & CutOut & 0.810\scriptsize$\pm$0.072 & 0.656\scriptsize$\pm$0.214 & 0.374\scriptsize$\pm$0.104 & 0.511\scriptsize$\pm$0.227 & 0.471\scriptsize$\pm$0.158 & 0.262\scriptsize$\pm$0.115 \\
 & CutMix & 0.733\scriptsize$\pm$0.107 & \underline{0.211\scriptsize$\pm$0.083} & \underline{0.097\scriptsize$\pm$0.017} &0.653\scriptsize$\pm$0.120 & 0.214\scriptsize$\pm$0.072 & \underline{0.082\scriptsize$\pm$0.031} \\
 & Aug & \underline{0.664\scriptsize$\pm$0.190} & 0.517\scriptsize$\pm$0.197 & 0.526\scriptsize$\pm$0.132 & 0.677\scriptsize$\pm$0.206 & \underline{0.115\scriptsize$\pm$0.053} & 0.317\scriptsize$\pm$0.081 \\
 & GDRO & 0.759\scriptsize$\pm$0.032 & 0.624\scriptsize$\pm$0.060 & 0.267\scriptsize$\pm$0.083 & 0.697\scriptsize$\pm$0.085 & 0.335\scriptsize$\pm$0.085 & 0.097\scriptsize$\pm$0.053 \\
 & JTT & 0.719\scriptsize$\pm$0.035 & 0.533\scriptsize$\pm$0.034 & 0.378\scriptsize$\pm$0.121 & 0.703\scriptsize$\pm$0.125 & 0.350\scriptsize$\pm$0.073 & 0.158\scriptsize$\pm$0.051 \\
 & Ours & \textbf{0.348\scriptsize$\pm$0.151}* & \textbf{0.106\scriptsize$\pm$0.051} & \textbf{0.059\scriptsize$\pm$0.044} & \textbf{0.077\scriptsize$\pm$0.109}* & \textbf{0.057\scriptsize$\pm$0.037} & \textbf{0.067\scriptsize$\pm$0.083} \\
 
\midrule 

\multirow{8}{*}{75}
 & ERM & 0.445\scriptsize$\pm$0.155 & 0.387\scriptsize$\pm$0.064 & 0.201\scriptsize$\pm$0.087 & \underline{0.253\scriptsize$\pm$0.054} & 0.273\scriptsize$\pm$0.057 & 0.186\scriptsize$\pm$0.077 \\
 & MixUp & \underline{0.380\scriptsize$\pm$0.120} & 0.331\scriptsize$\pm$0.067 & 0.156\scriptsize$\pm$0.036 & 0.470\scriptsize$\pm$0.149 & 0.282\scriptsize$\pm$0.095 & 0.183\scriptsize$\pm$0.036 \\
 & CutOut & 0.460\scriptsize$\pm$0.078 & 0.361\scriptsize$\pm$0.111 & 0.168\scriptsize$\pm$0.044 & 0.265\scriptsize$\pm$0.188 & \textbf{0.140\scriptsize$\pm$0.036} & 0.165\scriptsize$\pm$0.120 \\
 & CutMix & 0.526\scriptsize$\pm$0.095 & \underline{0.138\scriptsize$\pm$0.032} & \textbf{0.055\scriptsize$\pm$0.020} & 0.330\scriptsize$\pm$0.082 & 0.153\scriptsize$\pm$0.048 & 0.070\scriptsize$\pm$0.061 \\
 & Aug & 0.446\scriptsize$\pm$0.131 & 0.408\scriptsize$\pm$0.158 & 0.345\scriptsize$\pm$0.044 & 0.324\scriptsize$\pm$0.142 & 0.149\scriptsize$\pm$0.063 & 0.256\scriptsize$\pm$0.066 \\
 & GDRO & 0.507\scriptsize$\pm$0.019 & 0.329\scriptsize$\pm$0.015 & 0.114\scriptsize$\pm$0.024 & 0.444\scriptsize$\pm$0.064 & 0.183\scriptsize$\pm$0.023 & \textbf{0.057\scriptsize$\pm$0.010} \\
 & JTT & 0.452\scriptsize$\pm$0.046 & 0.261\scriptsize$\pm$0.035 & 0.146\scriptsize$\pm$0.048 & 0.459\scriptsize$\pm$0.077 & 0.204\scriptsize$\pm$0.029 & 0.086\scriptsize$\pm$0.037 \\
 & Ours & \textbf{0.364\scriptsize$\pm$0.182} & \textbf{0.127\scriptsize$\pm$0.059} & \underline{0.061\scriptsize$\pm$0.028} & \textbf{0.066\scriptsize$\pm$0.040}* & \underline{0.145\scriptsize$\pm$0.088} & \underline{0.063\scriptsize$\pm$0.024} \\
 \bottomrule
\\
\end{tabular}

\label{tab:tpr_disp_table_new}
\end{table*}

Having established the layer-specific nature of shortcut learning, we now demonstrate that our knowledge distillation framework, utilizing a teacher trained on a small curated dataset, effectively prevents students from developing shortcut dependencies across multiple datasets and bias types. 
We evaluate our approach across the CheXpert and ISIC datasets and all student models are trained on data augmented with a synthetic bias feature with different degrees of correlation with the class label (prevalence). 
We compare the performance of our model to several augmentation-based debiasing approaches, as well as Just Train Twice (JTT) and GroupDRO.

Across all tested bias types and degrees of correlation with class labels, our student network demonstrates the most consistently low bias in its predictions, as measured by $\Delta$TPR between bias-aligned and bias-contrasting samples (Table \ref{tab:tpr_disp_table_new}). 
In several cases, the TPR disparity is reduced such that it is comparable to our clean baseline.  
Our method remains similarly effective in reducing the bias even as its prevalence increases, while most other methods we evaluate worsen in effectiveness at higher prevalence rates. 
Notably, the majority of methods show significantly reduced efficacy in mitigating diffuse shortcuts across both datasets.
Contrastingly, our approach is consistently effective across localized and diffuse shortcuts. We highlight a consistent drop in efficacy for the noise bias feature in the CheXpert dataset with all methods, including ours. We suspect that this is due to useful textural information being corrupted by the noise shortcut. 
We also typically find that our approach achieves better overall AUC compared to other methods, particularly at higher prevalence rates. 
As shortcut prevalence decreases from 95\% to 75\%, all methods show improved $\Delta$TPR, which is expected since weaker biases will provide less misleading signal during training. 
However, even at lower prevalence rates, our approach maintains its advantage over other methods.

\begin{table*}[!th]
\centering
\caption{AUC $\uparrow$ for ResNet-18. We compare our approach to four popular augmentation-based de-biasing techniques. Shortcuts here have a 100\% correlation with the task label, so group-based methods (GroupDRO and JTT) are omitted from these comparisons. Results are presented as Mean$\pm$Std over 5-fold cross-validation. Models are marked as \textbf{best} and \underline{second-best}. When the difference between first and second best is statistically significant ($p<0.05$ according to a paired t-test), the best-performing model is marked *.}
\label{tab:auc_table}
\footnotesize
\begin{tabular}{c c c c c c c c}
\midrule
\multirow{2}{*}{Test set} &\multirow{2}{*}{Model}  & \multicolumn{3}{c}{CheXpert} & \multicolumn{3}{c}{ISIC} \\ \cmidrule(lr){3-5} \cmidrule{6-8}
 & & Noise & Square (C) & Square (R) & Noise & Square (C) & Square (R) \\
 \midrule
  \multirow{7}{*}{Biased}
 & Baseline & 0.709\scriptsize$\pm$0.024 & 0.755\scriptsize$\pm$0.013 & 0.752\scriptsize$\pm$0.015 & 0.749\scriptsize$\pm$0.024& 0.809\scriptsize$\pm$0.019 & 0.808\scriptsize$\pm$0.017  \\
 \cmidrule{2-8}
 & ERM & 0.489\scriptsize$\pm$0.012 & 0.533\scriptsize$\pm$0.007 & 0.554\scriptsize$\pm$0.006 & 0.521\scriptsize$\pm$0.011 & 0.600\scriptsize$\pm$0.011 & 0.612\scriptsize$\pm$0.007 \\
 & MixUp & 0.509\scriptsize$\pm$0.007 & 0.539\scriptsize$\pm$0.008 & 0.550\scriptsize$\pm$0.010 & 0.490\scriptsize$\pm$0.026 & 0.555\scriptsize$\pm$0.028 & 0.585\scriptsize$\pm$0.011 \\
 & CutOut  & 0.498\scriptsize$\pm$0.029 & 0.548\scriptsize$\pm$0.010 & 0.584\scriptsize$\pm$0.008 & \underline{0.521\scriptsize$\pm$0.023} & 0.627\scriptsize$\pm$0.013 & 0.639\scriptsize$\pm$0.006 \\
 & CutMix  & \underline{0.529\scriptsize$\pm$0.007} & \underline{0.680\scriptsize$\pm$0.025} & \underline{0.758\scriptsize$\pm$0.015} & 0.516\scriptsize$\pm$0.015 & \underline{0.753\scriptsize$\pm$0.038} & \underline{0.781\scriptsize$\pm$0.020} \\
 & Aug  & 0.483\scriptsize$\pm$0.009 & 0.585\scriptsize$\pm$0.017 & 0.550\scriptsize$\pm$0.013 & 0.518\scriptsize$\pm$0.006 & 0.731\scriptsize$\pm$0.017 & 0.631\scriptsize$\pm$0.010 \\
 & Ours & \textbf{0.689\scriptsize$\pm$0.044}* & \textbf{0.747\scriptsize$\pm$0.008}* & \textbf{0.761\scriptsize$\pm$0.01} & \textbf{0.775\scriptsize$\pm$0.023}* & \textbf{0.777\scriptsize$\pm$0.024} & \textbf{0.805\scriptsize$\pm$0.016} \\
 \midrule
 
\multirow{7}{*}{Clean}
 & Baseline & 0.754\scriptsize$\pm$0.014 & 0.754\scriptsize$\pm$0.014 & 0.754\scriptsize$\pm$0.014 & 0.811\scriptsize$\pm$0.019 & 0.811\scriptsize$\pm$0.019 & 0.811\scriptsize$\pm$0.019 \\
 \cmidrule{2-8}
 & ERM & 0.491\scriptsize$\pm$0.029 & 0.599\scriptsize$\pm$0.020 & 0.704\scriptsize$\pm$0.015 & \underline{0.498\scriptsize$\pm$0.038} & 0.672\scriptsize$\pm$0.038 & 0.745\scriptsize$\pm$0.015 \\
 & MixUp & 0.587\scriptsize$\pm$0.021 & 0.581\scriptsize$\pm$0.015  & 0.649\scriptsize$\pm$0.025 & 0.402\scriptsize$\pm$0.018 & 0.565\scriptsize$\pm$0.037 & 0.652\scriptsize$\pm$0.036 \\
 & CutOut & 0.527\scriptsize$\pm$0.063 & 0.604\scriptsize$\pm$0.012 & 0.741\scriptsize$\pm$0.007 & 0.485\scriptsize$\pm$0.089 & 0.722\scriptsize$\pm$0.021 & 0.758\scriptsize$\pm$0.013 \\
 & CutMix & \underline{0.608\scriptsize$\pm$0.017} & \underline{0.743\scriptsize$\pm$0.037} & \textbf{0.776\scriptsize$\pm$0.011} & 0.495\scriptsize$\pm$0.040 & \textbf{0.791\scriptsize$\pm$0.037} & \underline{0.797\scriptsize$\pm$0.016} \\
 & Aug  & 0.507\scriptsize$\pm$0.037 & 0.711\scriptsize$\pm$0.044 & 0.703\scriptsize$\pm$0.015 & 0.444\scriptsize$\pm$0.012 & 0.777\scriptsize$\pm$0.021 & 0.727\scriptsize$\pm$0.032 \\
 & Ours & \textbf{0.741\scriptsize$\pm$0.010}* & \textbf{0.749\scriptsize$\pm$0.009} & \underline{0.763\scriptsize$\pm$0.011} & \textbf{0.767\scriptsize$\pm$0.028}* & \underline{0.778\scriptsize$\pm$0.024} & \textbf{0.807\scriptsize$\pm$0.016}
 \\
 \midrule
 
 \multirow{7}{*}{OOD}
  & Baseline & 0.737\scriptsize$\pm$0.014 & 0.737\scriptsize$\pm$0.014 & 0.737\scriptsize$\pm$0.014 & 0.677\scriptsize$\pm$0.024 & 0.677\scriptsize$\pm$0.024 & 0.677\scriptsize$\pm$0.024 \\
 \cmidrule{2-8}
 & ERM & 0.461 $\pm$ 0.042 & 0.548\scriptsize$\pm$0.015 & 0.688\scriptsize$\pm$0.027 & 0.645\scriptsize$\pm$0.031 & 0.557\scriptsize$\pm$0.012 & 0.556\scriptsize$\pm$0.052 \\
 & MixUp & \underline{0.534\scriptsize$\pm$0.037} & 0.546\scriptsize$\pm$0.035 & 0.633\scriptsize$\pm$0.038 & 0.567\scriptsize$\pm$0.035 & 0.539\scriptsize$\pm$0.033 & 0.534\scriptsize$\pm$0.036 \\
 & CutOut & 0.424\scriptsize$\pm$0.066 & 0.571\scriptsize$\pm$0.049 & \underline{0.730\scriptsize$\pm$0.013}  & 0.635\scriptsize$\pm$0.016 & 0.585\scriptsize$\pm$0.008 & 0.583\scriptsize$\pm$0.020 \\
 & CutMix & 0.526\scriptsize$\pm$0.040 & \underline{0.692\scriptsize$\pm$0.013} & 0.724\scriptsize$\pm$0.021 & 0.650\scriptsize$\pm$0.020 & 0.561\scriptsize$\pm$0.034 & 0.529\scriptsize$\pm$0.043 \\
 & Aug & 0.426\scriptsize$\pm$0.027 & 0.683\scriptsize$\pm$0.018 & 0.692\scriptsize$\pm$0.008 & \underline{0.670\scriptsize$\pm$0.012} & \underline{0.635\scriptsize$\pm$0.028} & \underline{0.618\scriptsize$\pm$0.026} \\
 & Ours & \textbf{0.733\scriptsize$\pm$0.018}* & \textbf{0.759\scriptsize$\pm$0.022}* & \textbf{0.763\scriptsize$\pm$0.008}* & \textbf{0.727\scriptsize$\pm$0.057} & \textbf{0.697\scriptsize$\pm$0.030}*  & \textbf{0.666\scriptsize$\pm$0.042} \\
 \bottomrule
 \end{tabular}

\end{table*}

\subsubsection{Generalization to clean and out-of-distribution data}
A critical test of any deep neural network is whether it has learned a robust, generalizable set of decision rules. Here we evaluate our approach across three distinct evaluation scenarios: (1) a biased test set where shortcuts are present but distributed equally across classes, such that they are no longer useful predictive features; (2) a clean test set featuring none of the synthetic bias features present in our training sets; and (3) out-of-distribution (OOD) test sets to evaluate generalization. 
This allows us to assess both the method's ability to ignore spurious features and its capacity to learn robust, generalizable, and clinically relevant features. Our findings are highlighted in Table \ref{tab:auc_table}.

Comparisons on the bias-corrupted test set allow validation of how well each model learned to ignore the presence of the shortcuts at inference. 
Across all shortcut types on both datasets, we find that our method consistently achieves the best overall AUC and consistently matches or even outperforms the clean baseline evaluated on the shortcut-corrupted test data. 
We highlight that the clean baseline consistently sees a significant drop in performance when evaluated on noise-corrupted data. 
We hypothsize that this is likely a result of the degradation of useful texture-related information in the test set, combined with inherent bias of CNN architectures towards textural information \citep{geirhos2018imagenet}. 

Interestingly, we find that all models see significantly improved performance when tested on the clean test data. 
This supports previous findings that biases in the training data do not necessarily prevent models from learning underlying causal features \citep{stanley2025and, glocker2023risk}; but can lead them to preferentially rely on the spuriously correlated features when they are available. 
Notably, across all shortcut types, our model tested on the clean dataset achieves performance that is competitive with the baseline. 
By comparison, most other tested methods fail to see an improvement in AUC on the clean test set when the training data was augmented with the noise shortcut. 
We highlight this as further evidence of the power of a teacher model fine-tuned on a small amount of task-relevant data to prevent a student from being corrupted by the spurious feature. 

Finally, the OOD test sets serve to evaluate the robustness and generalizability of the decision rules learned by the network.
Here, we see that our student network consistently matches the performance of the clean baseline across both datasets and all bias features, consistently outperforming all other approaches. 

These findings collectively support our hypothesis that task-relevant knowledge distillation across intermediate network layers can effectively guide models toward learning more robust and clinically relevant features.

\subsubsection{Effectiveness against multiple concurrent shortcuts}
\begin{figure*}[h]
    \centering
    \includegraphics[width=\textwidth]{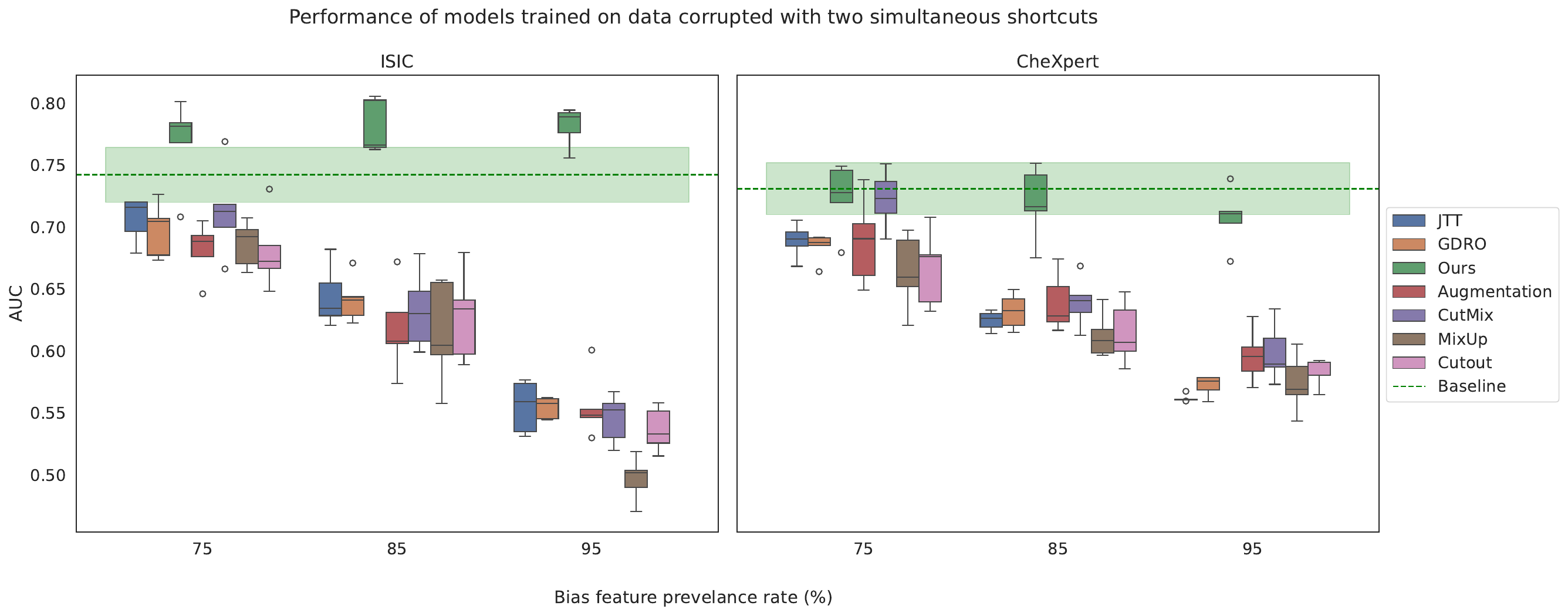}
    \caption{Performance of ResNet-18 trained on ISIC and CheXpert datasets featuring multiple simultaneous shortcuts. The green line represents a model trained on a training set before augmenting with synthetic shortcuts. We compare our student with a specialized teacher to JTT and GroupDRO.}\label{fig:multi_shortcut}
\end{figure*}

Our proposed approach has demonstrated promise in the mitigation of synthetic shortcuts. 
However, prior experiments purposefully represent a highly controlled shortcut environment. 
Only a single synthetic shortcut is present in the training data. 
Realistically, spurious features are unlikely to be constrained to a single source, particularly in large datasets. It is important, therefore, that any bias mitigation approach is able to mitigate multiple sources of bias simultaneously.

We augment our training data with two simultaneous shortcuts, one correlated with the positive class and the other with the negative class. 
The predictive strength of these shortcuts is varied across different training sets. 
As seen in Figure \ref{fig:multi_shortcut}, our method remains effective in the presence of multiple bias sources in the training data, and across all tested prevalence rates, consistently outperforming all other methods. 
In the majority of cases, we find that our student model remains competitive with the baseline model trained on entirely clean data.

\subsubsection{Validation on realistic 3D structural biases}
\begin{figure}[t]
    \centering
    \includegraphics[width=0.5\textwidth]{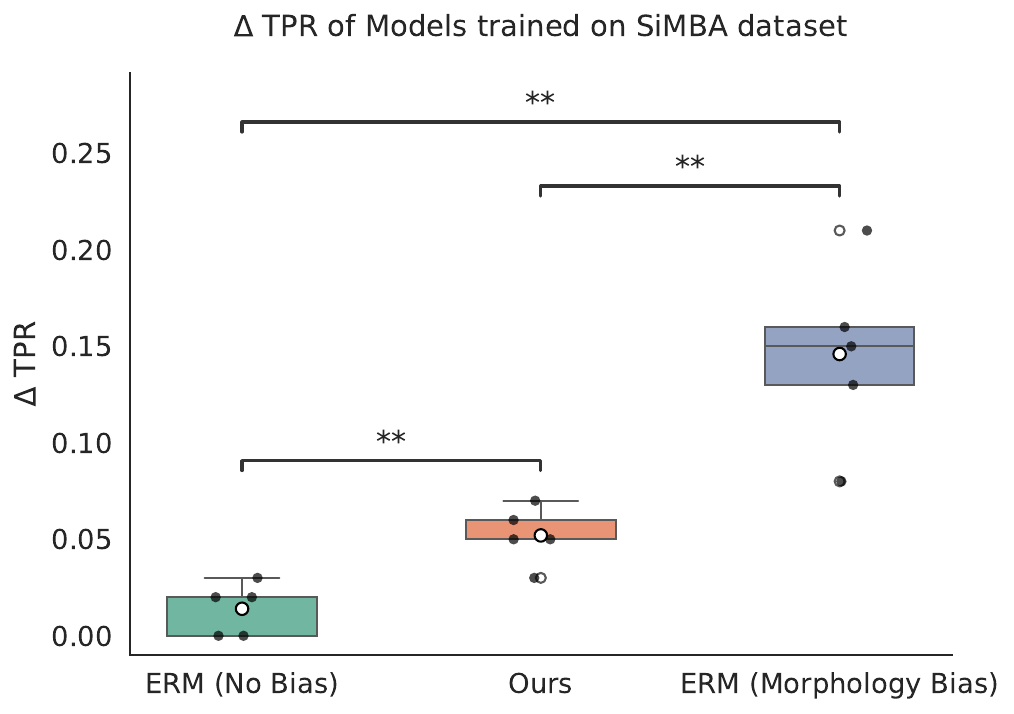}
    \caption{
    $\Delta$TPR of 3D CNN models trained on SimBA data. ERM (No Bias) is trained on data without any bias features. 
    Ours and ERM (Morphology Bias) are trained on data augmented with a synthetic morphological bias feature. 
    All models are evaluated on test data featuring the morphology bias.
    ** indicates a statistically significant difference in $\Delta$TPR according to a paired t-test with Bonferroni correction.
    }\label{fig:simba}
\end{figure}

While our previous experiments focused primarily on 2D image classification with synthetic shortcuts, we now extend our analysis to a more realistic scenario featuring subtle structural biases that more closely resemble real-world medical imaging artifacts. To evaluate our approach in this realistic context, we leverage the SimBA dataset — a synthetic brain MRI dataset designed specifically to study bias in 3D medical image analysis \citep{stanley2024towards}. The SimBA dataset features subtle morphological deformations that correlate with disease labels at a 65\% prevalence rate. These localized structural modifications represent a more nuanced and challenging form of bias compared to our previous experiments with artificial shortcuts.

Importantly, a key methodological distinction in these experiments is that all data splits in SimBA (training, validation, and test) exhibit the same bias prevalence rate. This differs from our previous synthetic shortcut experiments, where validation data contained balanced shortcut distributions. The absence of bias-balanced validation data creates a significantly more challenging scenario that closely mirrors real-world clinical settings, where validation data often shares the same biases as training data.

We train a lightweight 3D CNN with linear classification probes attached after each convolutional layer. Our teacher model is trained on a 20\% subset of the unbiased training data, while the student model is trained on the biased dataset. All volumes were resampled to 96×96×96 voxels.

Figure \ref{fig:simba} presents the performance comparison between three models: our student model guided by a teacher fine-tuned on some task-relevant data (Ours), a model trained on the full unbiased dataset (ERM (No Bias)), and a model trained on the biased dataset using standard Empirical Risk Minimization (ERM (Morphology Bias)). The results demonstrate clear performance differences among these approaches.

Statistical analysis using repeated measures ANOVA confirms significant differences between the models. Subsequent pairwise comparisons using paired t-tests with Bonferroni correction reveal statistically significant differences in $\Delta$TPR between the ERM model trained on the morphologically biased dataset and both alternative models. Notably, while a statistically significant difference in $\Delta$TPR remains between our model and ERM (No Bias), the disparity difference is significantly reduced compared to ERM (Morphology Bias). 

Our findings demonstrate that our approach can effectively mitigate bias even without the benefit of a balanced validation set to guide the training process. This is significant for real-world medical imaging applications, where obtaining bias-balanced validation data is often infeasible. These results further validate the applicability of our method to complex 3D medical imaging tasks featuring realistic bias patterns, suggesting broader potential for clinical applications.

\subsection{Method design and optimization}

Having validated our core approach, we now explore key design choices that optimize its effectiveness and practical applicability.

\subsubsection{Partial layer distillation preserves student performance}\label{sec:partial}

\begin{table*}[t]
\centering
\footnotesize
\caption{Performance of a ResNet-18 student network tested on the shortcut-corrupted test sets with our distillation loss. Distillation loss is applied at different numbers of intermediate layers between 0 and 17. When loss is applied at 0 intermediate layers, we only apply KD between the student and teacher's final outputs.  Results are presented as Mean$\pm$Std over 5-fold cross-validation. Models are marked \textbf{best} and \underline{second-best}.}
\begin{tabular}{c c c c c c c c c}
 \midrule
\multirow{2}{*}{} & \multirow{2}{*}{\# layers} & \multicolumn{3}{c}{CheXpert} & \multicolumn{3}{c}{ISIC} \\ \cmidrule(lr){3-5} \cmidrule{6-8}
& & Noise & Square (C) & Square (R) & Noise & Square (C) & Square (R) \\ \midrule
\multirow{5}{*}{AUC $\uparrow$}
& 17 & \textbf{0.694\scriptsize$\pm$0.034} & 0.742\scriptsize$\pm$0.009 & 0.762\scriptsize$\pm$0.008 & 0.754\scriptsize$\pm$0.019 & 0.761\scriptsize$\pm$0.035 & 0.754\scriptsize$\pm$0.023\\
& 13 & 0.688\scriptsize$\pm$0.034 & 0.746\scriptsize$\pm$0.007 & \underline{0.762\scriptsize$\pm$0.005} & \textbf{0.780\scriptsize$\pm$0.018} & \underline{0.767\scriptsize$\pm$0.012} & 0.780\scriptsize$\pm$0.011\\ 
& 9 & 0.687\scriptsize$\pm$0.034 & \underline{0.747\scriptsize$\pm$0.008} & 0.756\scriptsize$\pm$0.014 & 0.768\scriptsize$\pm$0.013 & 0.762\scriptsize$\pm$0.033 & \underline{0.783\scriptsize$\pm$0.015}\\ 
& 5 & \underline{0.689\scriptsize$\pm$0.044} & \textbf{0.747\scriptsize$\pm$0.008} & \textbf{0.762\scriptsize$\pm$0.010} & \underline{0.775\scriptsize$\pm$0.023} & \textbf{0.777\scriptsize$\pm$0.024} & \textbf{0.807\scriptsize$\pm$0.016}\\  
&  0 & 0.606\scriptsize$\pm$0.008 & 0.640\scriptsize$\pm$0.015 & 0.684\scriptsize$\pm$0.014 & 0.632\scriptsize$\pm$0.019 & 0.667\scriptsize$\pm$0.013 & 0.713\scriptsize$\pm$0.017 \\
\midrule
\multirow{5}{*}{$\Delta$TPR $\downarrow$}
& 17 & \textbf{0.272\scriptsize$\pm$0.07} & \underline{0.083\scriptsize$\pm$0.024} & \textbf{0.028\scriptsize$\pm$0.020} & 0.091\scriptsize$\pm$0.037 & 0.041\scriptsize$\pm$0.038 & \textbf{0.028\scriptsize$\pm$0.02}\\
& 13 & 0.378\scriptsize$\pm$0.188 & 0.107\scriptsize$\pm$0.018 & 0.049\scriptsize$\pm$0.030 & \underline{0.074\scriptsize$\pm$0.034} & \textbf{0.019\scriptsize$\pm$0.021} & 0.049\scriptsize$\pm$0.030\\ 
& 9 & \underline{0.301\scriptsize$\pm$0.115} & 0.083\scriptsize$\pm$0.053 & 0.046\scriptsize$\pm$0.023 & 0.077\scriptsize$\pm$0.027 & 0.054\scriptsize$\pm$0.033 & 0.046\scriptsize$\pm$0.023\\ 
& 5 & 0.377\scriptsize$\pm$0.185 & \textbf{0.079\scriptsize$\pm$0.017} & \underline{0.032\scriptsize$\pm$0.020} & \textbf{0.068\scriptsize$\pm$0.055} & \underline{0.038\scriptsize$\pm$0.032} & 0.032\scriptsize$\pm$0.020\\  
& 0 & 0.831\scriptsize$\pm$0.091 & 0.662\scriptsize$\pm$0.126 & 0.424\scriptsize$\pm$0.074 & 0.813\scriptsize$\pm$0.139 & 0.617\scriptsize$\pm$0.145 & 0.424\scriptsize$\pm$0.074\\
 \bottomrule
  \end{tabular}
\label{tab:fewer_probes_tpr_disp_table}
\end{table*}

While our initial implementation applied knowledge distillation across all batch normalization layers of our ResNet-18 students, this comprehensive approach might over-constrain the students' learning process. 
We investigate whether more selective application of distillation can maintain or even enhance performance.
We systematically evaluate distillation applied at varying numbers of intermediate layers in a ResNet-18 network, from all 17 intermediate layers to 0 intermediate layers (final classification head only).

For partial-layer configurations, we employ a random sampling approach where we independently select \( n \) layers from both the student and teacher networks during each training epoch. Importantly, these selections are made independently, meaning the specific layers chosen may differ between networks. We pair the selected layers sequentially based on their relative depth to establish meaningful knowledge transfer despite potentially different architectural positions.

Table \ref{tab:fewer_probes_tpr_disp_table} reveals key insights about the value of our intermediate-layer distillation. 
We note that applying distillation at fewer intermediate layers ($5 - 9$) leads to comparable performance to distillation applied at all intermediate layers ($17$), both in terms of AUC and $\Delta$TPR. 
In some cases on the ISIC dataset, we see improvements in the AUC when the loss is applied at fewer layers. 
We hypothesize that in these cases, the reduced regularization of the KD loss facilitates an improved ability of the network to learn task-relevant features without sacrificing the useful guidance away from spurious features.
Importantly, when distillation is applied solely at the final classification head and not in the intermediate layers ($n = 0$), performance declines significantly and bias increases significantly across all experiments. 
This dramatic deterioration highlights the critical role of intermediate-layer guidance in mitigating shortcut learning.

\begin{table}[t]
\centering
\caption{AUC$\uparrow$ of a ResNet-18 and DenseNet-121 trained and evaluated on shortcut-corrupted data. We compare student models trained following our knowledge distillation protocol using a low-capacity teacher network (Ours) to models trained following standard cross-entropy optimization (ERM). The best-performing model is in \textbf{bold}.}
\footnotesize
\begin{tabular}{c c c c c c c}
\midrule
\multirow{2}{*}{} & \multirow{2}{*}{} & \multicolumn{3}{c}{CheXpert} \\ \cmidrule{3-5}
& & Noise & Square (C) & Square (R)  \\
 \midrule
 
\multirow{2}{*}{ResNet-18} 
& Ours  & \textbf{0.68\scriptsize±0.01} & \textbf{0.74\scriptsize±0.02} & \textbf{0.75\scriptsize±0.02}\\
 & ERM & 0.49\scriptsize±0.01 & 0.53\scriptsize±0.01 & 0.55\scriptsize±0.01 \\
\\
\multirow{2}{*}{DenseNet-121}  
& Ours & \textbf{0.63\scriptsize±0.02} & \textbf{0.69\scriptsize±0.01} & \textbf{0.70\scriptsize±0.02} \\
 & ERM & 0.50\scriptsize±0.01 & 0.52\scriptsize±0.01 & 0.53\scriptsize±0.01 & \\
 \midrule
 
 \multirow{2}{*}{} & \multirow{2}{*}{} & \multicolumn{3}{c}{ISIC} \\ \cmidrule{3-5}
 & & Noise & Square (C) & Square (R) \\
 \midrule
 
 \multirow{2}{*}{ResNet-18} 
& Ours  & \textbf{0.76\scriptsize±0.02} & \textbf{0.77\scriptsize±0.03} & \textbf{0.77\scriptsize±0.03}\\
 & ERM & 0.52\scriptsize±0.01 & 0.60\scriptsize±0.01 & 0.61\scriptsize±0.01 \\
\\
\multirow{2}{*}{DenseNet-121}  
& Ours & \textbf{0.72\scriptsize±0.01} & \textbf{0.72\scriptsize±0.03} & \textbf{0.74\scriptsize±0.03} \\
 & ERM & 0.51\scriptsize±0.01&0.63\scriptsize±0.01 & 0.62\scriptsize±0.01 \\
 \bottomrule
\end{tabular}
\label{tab:low_capacity_teacher}
\end{table}

\subsubsection{Low-capacity teachers effectively guide larger student networks}

Training the teacher using a small, curated subset of data can pose challenges when applied to significantly larger models. 
In this study, we examine whether a low-capacity model can effectively serve as a teacher for a higher-capacity student. 
Specifically, we distill knowledge from an AlexNet teacher to a ResNet-18 student, and from a ResNet-18 teacher to a DenseNet-121 student. 
To apply knowledge distillation from a low-capacity teacher, we follow a similar protocol to Section \ref{sec:partial}. 
We randomly sample $n$ layers from the student network each epoch, where $n$ is equal to the number of classification probes in the teacher model. 
The $n$ layers of the student network are paired sequentially with the classification probes of the teacher. 

We train our student on our datasets augmented with synthetic biases and present these results in Table \ref{tab:low_capacity_teacher}. We compare our student to an identical network trained following a standard Cross Entropy optimization protocol (ERM). 
Even with a small teacher network, knowledge distillation from the intermediate layers proves capable of effectively mitigating the influence of shortcuts present in the student training data. 

 In real-world applications, it is more likely that larger models, such as DenseNet-121—often considered state-of-the-art—are employed instead of smaller networks such as a ResNet-18. 
Training a much larger teacher network on a very limited clean subset increases the likelihood that the teacher will overfit to its training data, negatively impacting its ability to guide the student network toward robust and generalizable features. 
The efficacy of compact teacher networks is, therefore, significant for the practical implementation of our approach.

\subsubsection{Task-specific teacher fine-tuning outperforms alternative approaches}
\begin{figure*}[t]
\centering
\includegraphics[width=\textwidth]{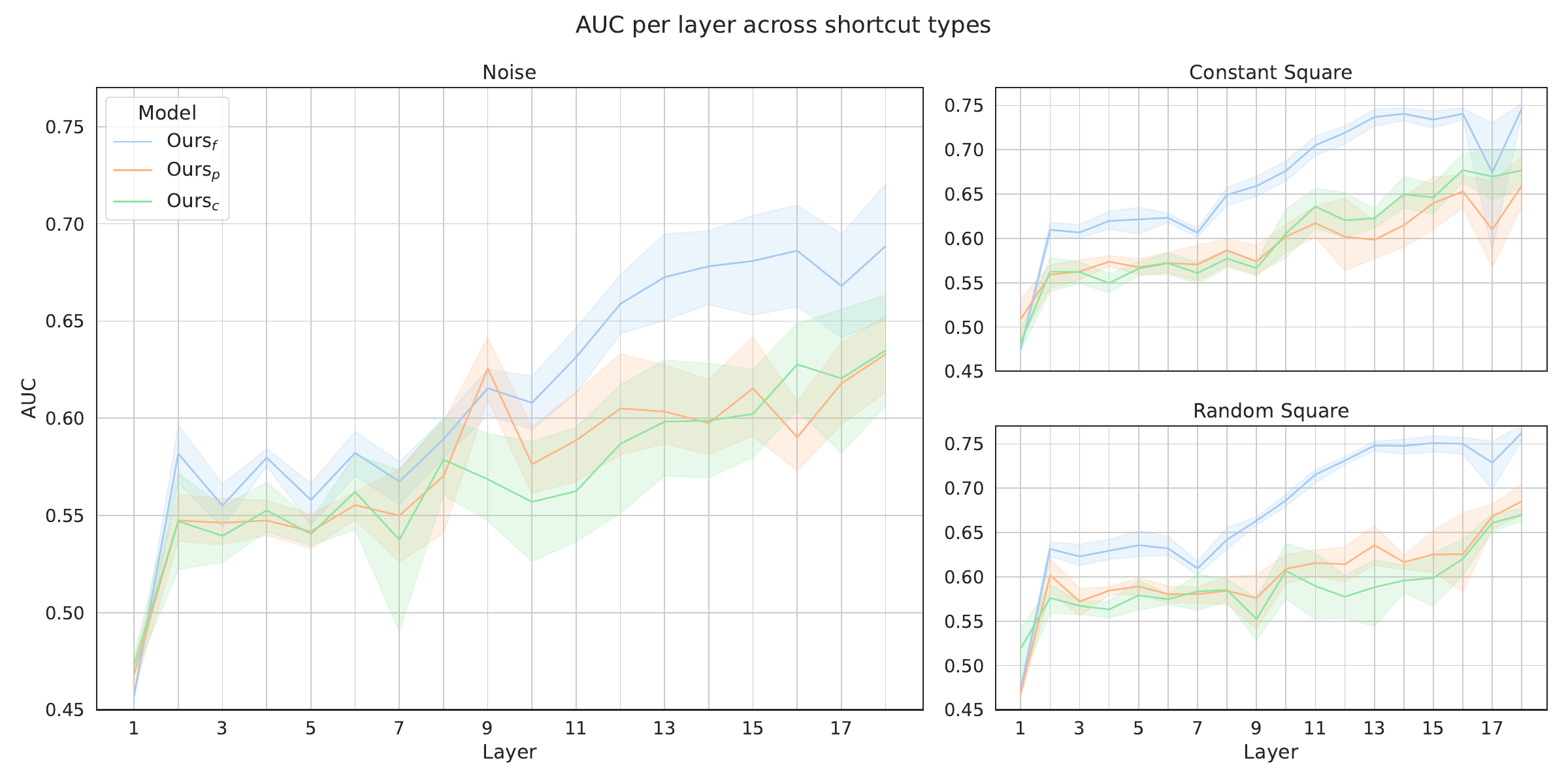}
\caption{Per-layer AUC of ResNet-18 students trained on CheXpert data featuring various synthetic shortcuts. Ours$_{f}$ is our fine-tuned teacher model, Ours$_{p}$ uses an ImageNet-pretrained ResNet-18 as a teacher, and Ours$_{c}$ applies pure confidence regularization in the intermediate layers.}\label{fig:teacher-per-layer-comp}
\end{figure*}

\begin{table*}[!th]
\centering
\footnotesize
\caption{Performance of student models with different knowledge distillation approaches. Ours$_{f}$ is our fine-tuned teacher model, Ours$_{p}$ uses an ImageNet-pretrained ResNet-18 as a teacher, and Ours$_{c}$ applies pure confidence regularization in the intermediate layers. Results are presented as Mean$\pm$Std over 5-fold cross-validation. Models are marked as \textbf{best} and \underline{second-best}. When the difference between first and second best is statistically significant, the best-performing model is highlighted with *.}
\begin{tabular}{c c c c c c c c c}
\midrule
&\multirow{2}{*}{Model} & \multicolumn{3}{c}{CheXpert} & \multicolumn{3}{c}{ISIC} & \\
& & Noise & Square (C) & Square (R) & Noise & Square (C) & Square (R) \\
 \midrule
\multirow{3}{*}{AUC $\uparrow$}
& Ours$_{f}$ & \textbf{0.689\scriptsize$\pm$0.044} & \textbf{0.747\scriptsize$\pm$0.008}* & \textbf{0.763\scriptsize$\pm$0.010}* & \textbf{0.775\scriptsize$\pm$0.023} & \textbf{0.777\scriptsize$\pm$0.024}* & \textbf{0.807\scriptsize$\pm$0.016}  \\
& Ours$_{p}$ & 0.633\scriptsize$\pm$0.025 & 0.660\scriptsize$\pm$0.039 & \underline{0.684\scriptsize$\pm$0.025} & 0.727\scriptsize$\pm$0.034 & 0.680\scriptsize$\pm$0.029 & \underline{0.782\scriptsize$\pm$0.016} \\
& Ours$_{c}$ & \underline{0.635\scriptsize$\pm$0.037} & \underline{0.677\scriptsize$\pm$0.029} & 0.673\scriptsize$\pm$0.007 & \underline{0.753\scriptsize$\pm$0.019} & \underline{0.731\scriptsize$\pm$0.023} & 0.696\scriptsize$\pm$0.071  \\
\midrule
\multirow{3}{*}{$\Delta$TPR $\downarrow$}
& Ours$_{f}$ & 0.377\scriptsize$\pm$0.185 & \textbf{0.079\scriptsize$\pm$0.017} & \textbf{0.034\scriptsize$\pm$0.016} & \underline{0.068\scriptsize$\pm$0.055} & \textbf{0.038\scriptsize$\pm$0.032} & \textbf{0.070\scriptsize$\pm$0.043} \\
& Ours$_{p}$ & \textbf{0.285\scriptsize$\pm$0.082} & 0.239\scriptsize$\pm$0.196  & \underline{0.115\scriptsize$\pm$0.094} & 0.218\scriptsize$\pm$0.123 & 0.318\scriptsize$\pm$0.162 & \underline{0.084\scriptsize$\pm$0.057} \\
& Ours$_{c}$ & \underline{0.351\scriptsize$\pm$0.073} & \underline{0.109\scriptsize$\pm$0.089}  & 0.140\scriptsize$\pm$0.062 & \textbf{0.036\scriptsize$\pm$0.033} & \underline{0.070\scriptsize$\pm$0.067} & 0.197\scriptsize$\pm$0.172 \\
\bottomrule
\end{tabular}
\label{tab:teacher_comparison}
\end{table*}
We propose that a teacher network fine-tuned on a small subset of task-relevant data can provide sufficient insight to deter a student network from learning bias features. 
Here, we validate this choice. We consider two alternative approaches to our proposed fine-tuned teacher to evaluate the importance of task-specific knowledge in the teacher: 

\begin{enumerate}
\item \textbf{ImageNet pre-trained teacher:} We use a teacher network pre-trained on the ImageNet dataset without any task-specific fine-tuning. 
This teacher possesses general visual recognition capabilities from training on diverse natural images but lacks task-specific or domain-specific medical imaging knowledge. 
Knowledge distillation is performed identically as with our fine-tuned teacher, with KL divergence minimization between corresponding intermediate layers of the student and the pre-trained teacher. 
This comparison helps us understand whether general visual features from a diverse dataset are sufficient for guiding the student away from shortcuts or if task-specific knowledge is essential.

\item \textbf{Confidence Regularization:} The teacher model is removed entirely in favor of a form of self-regularization. 
Rather than distilling knowledge from a teacher, we encourage the student network to maintain low confidence in its intermediate layer predictions by minimizing the KL divergence between each layer's predictions and a uniform class probability distribution. 
This forces the model to avoid becoming overconfident in any particular features too early in the network, potentially discouraging reliance on simple shortcut features. 
By comparing against this approach, we can determine whether the specific guidance from a teacher model provides advantages beyond simply preventing early layer overconfidence.
\end{enumerate}
Both alternatives represent reasonable approaches towards mitigating shortcut learning: the ImageNet teacher by transferring robust general visual representations, and confidence regularization by directly discouraging overconfidence in features at any particular layer. 
Our fine-tuned specialist teacher consistently outperforms both alternatives across most datasets and shortcut types, as shown in Table \ref{tab:teacher_comparison}. This performance is achieved without sacrificing fairness. This highlights that a teacher with task-specific knowledge is better equipped to guide the student away from simplistic shortcut features and toward more robust, task-relevant features. This is further supported by our findings in Figure \ref{fig:teacher-per-layer-comp}, which shows that our student trained with a fine-tuned teacher network achieves consistently higher AUC in the intermediate layers, and that the AUC of our student reaches higher levels earlier in the model.

\begin{figure*}[t]
    \centering
    \includegraphics[width=\textwidth]{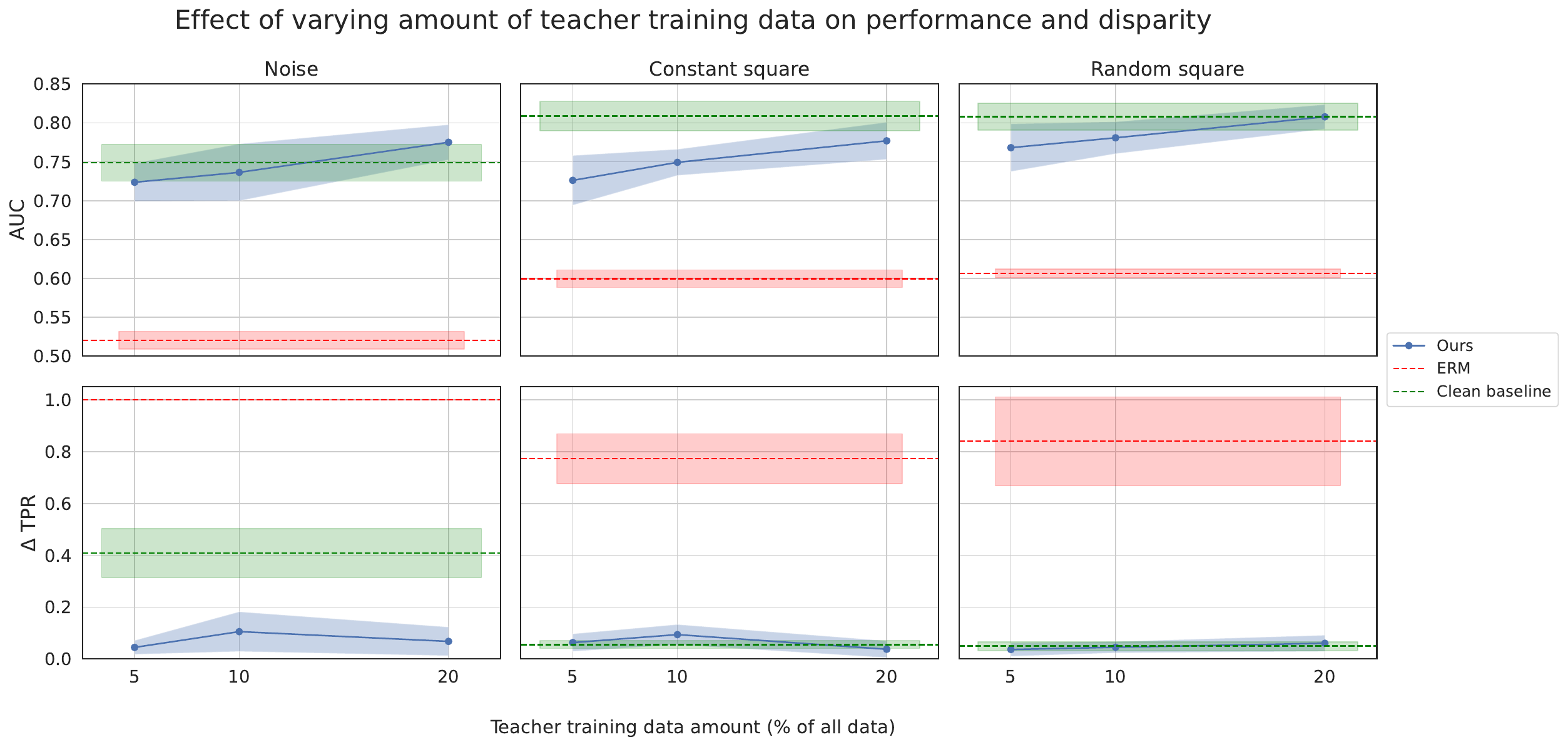}
    \caption{AUC $\uparrow$ (top) and $\Delta$TPR $\downarrow$ (bottom) of ResNet-18 students trained on ISIC data featuring various synthetic shortcuts. We vary the proportion of the original training data used to train the teacher network, using subsets consisting of 5\%, 10\%, and 20\% of the original training data. In each case, teacher training data is excluded from the student's training. All shortcuts have a 100\% prevalence in student training data. As shortcut reliance increases, overall performance (AUC) declines and performance disparity ($\Delta$TPR) increases.}\label{fig:teacher_training_data}
\end{figure*}

\subsection{Practical considerations for teacher model training}
Finally, we address critical practical questions about teacher model requirements that determine the real-world viability of our approach.

\subsubsection{Teacher effectiveness scales with training data volume}
A critical question for practical implementation is the volume of unbiased data required to train an effective teacher model. 
In our previous experiments, we evaluated our approach using a teacher network trained on 20\% of the full training data from each dataset. 
To better understand the amount of required data, we now assess the efficacy of our approach when the teacher network is trained on as little as 5\% and 10\% of the total training data.
In each case, the teacher training data is excluded from the student's training. As a result, each student network is trained on 95\%, 90\%, and 80\% of the full training data, depending on the amount of data used to train the teacher. 

As illustrated in Figure \ref{fig:teacher_training_data}, we observe a clear relationship between teacher training data volume and student performance. 
Bias metrics and overall model performance both improve consistently as the amount of training data used for the teacher increases. 
Notably, even when our teacher network is trained on as little as 5\% of our original training data (56 images for ISIC), we still observe a substantial reduction in bias compared to ERM training.

The consistent performance advantage observed with minimal unbiased data has significant implications for real-world applications. In clinical settings, where it is often challenging to obtain large amounts of bias-free data, our results indicate that even a small, carefully curated dataset can effectively guide the mitigation of shortcut learning. This finding greatly enhances the practical applicability of our approach, making it more feasible in resource-constrained environments where extensive manual annotation or bias identification could be prohibitively expensive. Although the need to curate an unbiased training set is not completely eliminated, the amount of teacher training data required may be modest enough to be achievable in many practical scenarios.

\subsubsection{Leveraging OOD data for teacher training maintains effectiveness}
\begin{figure}[t]
    \centering
    \includegraphics[width=0.5\textwidth]{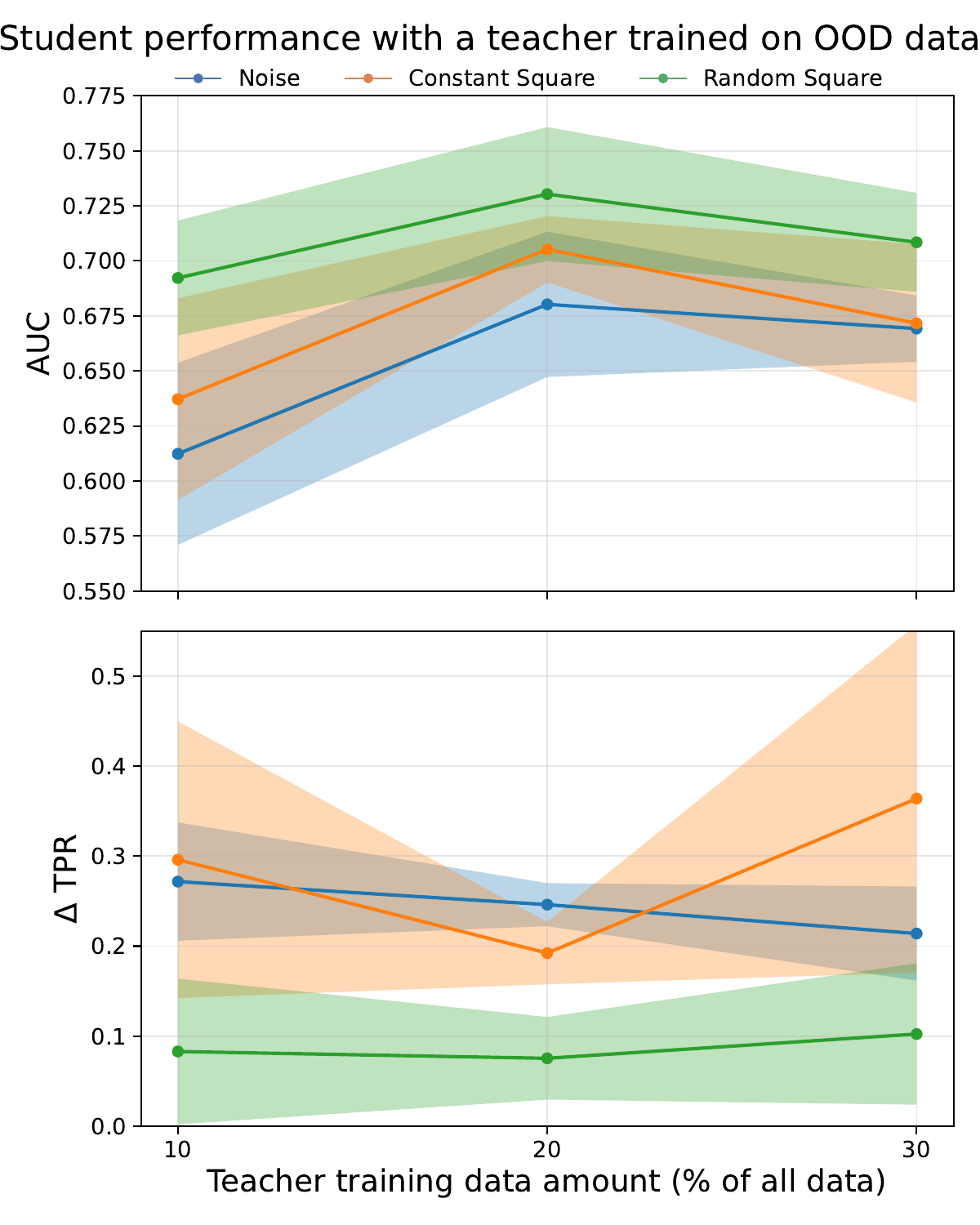}
    \caption{
    $\Delta$TPR and AUC of a ResNet-34 student trained on CheXpert with various synthetic shortcuts. Teacher model is trained on MIMIC at various subset sizes (between 10\% - 30\%). The student is trained on the full CheXpert training split with a shortcut prevalence of 95\%.
    }\label{fig:ood_teacher}
\end{figure}

In practice, obtaining curated teacher training data from the same distribution as the student's may not always be feasible. We investigate whether teacher models whose training data is OOD from the student's can still effectively guide bias mitigation. For this, we focus on the task of pneumothorax detection, training the teacher on MIMIC while the student is trained on CheXpert: both chest X-ray datasets, but from different institutions.

Figure \ref{fig:ood_teacher} demonstrates that our approach remains effective when teacher training data is OOD relative to the student. Performance improvements scale with teacher data volume, though OOD teachers require substantially more training data than in-distribution teachers. For example, our teacher trained on 10\% of the MIMIC training split is trained on approximately 400 images. By comparison, we see superior performance in a student trained with an in-distribution CheXpert teacher trained on 10\% of the CheXpert train split (approximately 140 images).
This increased data requirement likely reflects the underlying distribution shift between the datasets and the requirement for the teacher network to have learned robust features that transfer across institutional differences in imaging protocols and patient populations.
We also observe that ResNet-18 teachers struggle on OOD test sets when they have been trained on very little data, while ResNet-34 models perform better under the same circumstances. This suggests that the increased model capacity facilitates learning more generalizable features, particularly on smaller training sets.

These findings enhance practical applicability by demonstrating that OOD training data can be used to train the teacher model where it is not possible to curate bias-free in-distribution data. However, when using OOD data to train the teacher it is important to consider to larger data requirements required to achieve comparable performance.

\subsubsection{Robustness to shortcut features in teacher training data}\label{sec:shortcut_teacher}

A fundamental assumption of our work up until this point is the availability of perfectly clean training data for our teacher, free of all shortcuts present in the student's training data. However, this assumption may be unrealistic in practice, where subtle biases, such as demographic features or complex acquisition artifacts, can interact in unexpected ways that make the identification and removal of all shortcut features extremely challenging or impossible. To address this limitation, we investigate the robustness of our approach when the teacher's training data contains residual shortcut features at low prevalence rates.

We evaluate scenarios where shortcut features appear in 5\%, 10\%, and 15\% of positive-class samples (and in no negative-class samples) in the teacher's training data, while maintaining much higher prevalence in the student's training data. This simulates realistic conditions where shortcut learning mitigation efforts may not be able to guarantee, even with smaller, manually curated training sets, that the teacher's training data is entirely bias-free. We focus our analysis on the CheXpert dataset, training ResNet-18 models using the same protocol as in Section \ref{sec:kd_main_res}.

\begin{figure*}[t]
    \centering
    \includegraphics[width=\textwidth]{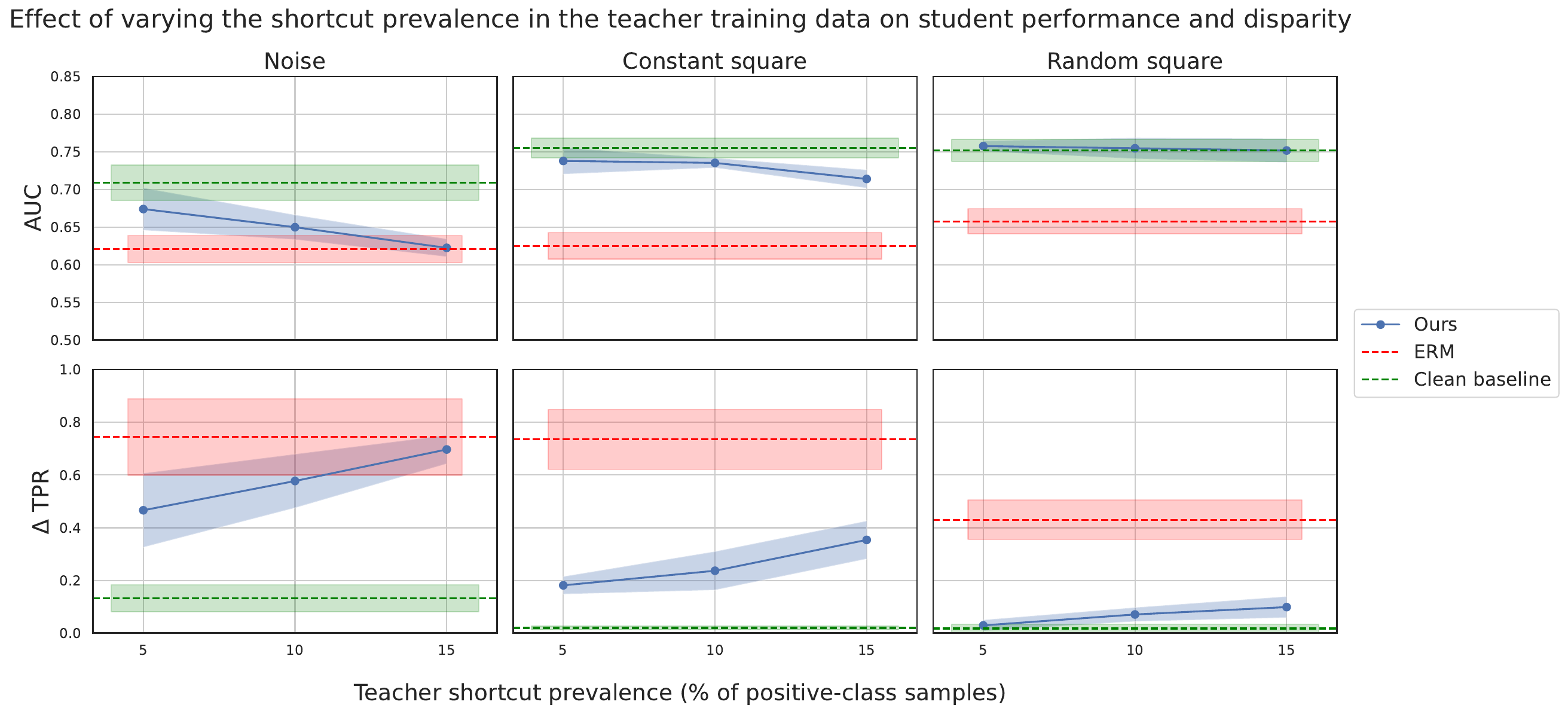}
    \caption{AUC $\uparrow$ (top) and $\Delta$TPR $\downarrow$ (bottom) of ResNet-18 students trained on CheXpert data featuring various synthetic shortcuts. We vary the prevalence of the shortcut in the data used to train the teacher network. In each case, the teacher is trained on a subset of 20\% of the full training split, and the student is trained on the remaining 80\%. The shortcut feature has a prevalence of 85\% in the student's training data across all experiments. As shortcut reliance increases, overall performance (AUC) declines and performance disparity ($\Delta$TPR) increases.}\label{fig:student_with_shortcut_teacher}
\end{figure*}

The visual complexity of the shortcut has a material impact on the prevalence at which we begin to observe disparities in performance. Figure \ref{fig:student_with_shortcut_teacher} illustrates both overall performance (AUC) and disparity ($\Delta$TPR) of a student model as the prevalence of shortcut features in the teacher's training data increases. For complex shortcuts like the random square pattern, even with 15\% prevalence in teacher data, both AUC and $\Delta$TPR of the student remain comparable with the clean baseline. In contrast, simpler shortcuts (noise and constant square) show greater sensitivity to teacher data contamination, with noticeable degradation even at a prevalence of 5\%. Such findings illustrate that very simple shortcut features significantly influence model learning even at very low prevalence in the training data.

Our findings align with the concept of ``availability'' introduced by \cite{hermann2023foundations}, who demonstrate that deep learning model's preferentially utilize the most \textit{available} features of their training data (i.e., those which are most easily identifiable), even if they are less \textit{predictive} than more challenging features. The greater availability of our low-level, simpler shortcut features (noise and constant square) compared to the random square shortcut or any disease feature leads the network to rely more heavily on these features, even if they are present in as little as 5\% of positive-class samples.

While the curation of bias-free teacher training data remains ideal, where the identification and removal of all possible shortcuts may be impossible or prohibitively time-consuming and costly, teacher dataset curation should focus on identifying and removing the most easily identifiable shortcut features (e.g., treatment devices, hospital logos, obvious markings). Prioritizing these most available features provides the greatest benefit for teacher effectiveness.


\section*{Discussion}
This paper addresses the critical challenge of shortcut learning in medical image analysis, proposing a novel knowledge distillation method leveraging teacher models fine-tuned on a small amount of unbiased, task-relevant data to guide student models towards robust features of their training data and away from bias features. 
Our findings highlight several key insights and practical advancements:

First, we demonstrate that shortcut learning manifests as distinct patterns of overconfidence at intermediate network layers, dependent on the type of shortcut involved. 
Diffuse shortcuts, such as noise patterns, tend to emerge in earlier network layers, suggesting that they do not require significant disambiguation to identify.
In contrast, localized shortcuts like geometric shapes manifest in later layers, indicating they require more complex feature disambiguation  (Figure \ref{fig:per_layer_conf_of_shortcuts}) . 

This layer-specific manifestation has important implications for both shortcut detection and mitigation. 
The early appearance of diffuse shortcuts suggests that initial network layers are particularly susceptible to learning simple, texture-related spurious correlations. 
This aligns with previous findings about the hierarchical nature of neural network learning, where early layers typically learn basic features while deeper layers capture more complex patterns \citep{baldock2021deep, chen2020towards}. 
The observation that different shortcuts manifest at different depths suggests that effective mitigation strategies should consider the network's entire processing pipeline rather than focusing solely on the final classification layer.

This is supported by our finding that distillation from an unbiased teacher to the intermediate layers of a student more effectively mitigates shortcut learning than distillation based solely on the final output (Table \ref{tab:fewer_probes_tpr_disp_table}). 

This finding offers a more nuanced understanding of how unwanted correlations manifest within the network's internal representations, and we believe that these insights are valuable beyond the specific method we propose here.
For example, such an observation may serve as an effective tool to monitor the learning and performance of deep neural networks to identify when they may be relying on easy spurious features.

A key contribution of our work is demonstrating that knowledge distillation from a teacher network trained on a small curated dataset significantly outperforms traditional de-biasing approaches  (Tables \ref{tab:tpr_disp_table_new} \& \ref{tab:auc_table}). 
Our method effectively prevents the student network from learning to rely on bias features present in their training data, surpassing traditional empirical risk minimization and alternative approaches such as confidence regularization or using ImageNet-pretrained teachers (Table \ref{tab:teacher_comparison}).
The approach consistently improves generalization and robustness, evidenced by substantial performance gains on both in-distribution and out-of-distribution test sets for the CheXpert and ISIC datasets (Table \ref{tab:auc_table}).

Our results demonstrate that selective intermediate-layer distillation can be as effective as comprehensive distillation across all network layers. 
As shown in Table \ref{tab:fewer_probes_tpr_disp_table}, distilling knowledge at only 5-9 layers consistently achieved comparable or superior performance to full 17-layer distillation, both in terms of AUC and $\Delta$TPR. 
This finding suggests that comprehensive distillation across all layers may be unnecessary in most cases and could even add excessive regularization to the student network's learning. 
While our random layer sampling approach proved effective, it represents a naive strategy that does not consider layer-specific contributions to shortcut learning. 
Future work should explore principled methods for identifying the layers where distillation would be most impactful. 
A more targeted distillation approach could further improve the effectiveness of mitigating shortcut learning.


Importantly, we demonstrate that compact architectures, such as AlexNet, can effectively guide larger, more sophisticated networks (ResNet-18 and DenseNet-121), addressing practical constraints related to training high-capacity models on small, unbiased datasets (Table \ref{tab:low_capacity_teacher}). 
This finding is critical for practical deployment in clinical contexts, where limited availability of unbiased data and computational constraints can limit the use of larger, resource-intensive models.

While our experiments are restricted to CNN-based architectures, transformer architectures are increasingly prevalent in medical image analysis literature. 
Many KD methods designed for CNNs that leverage the feature-space representations are not directly applicable to transformer networks due to the architectural differences. 
We suggest that since we do not leverage feature vectors directly, our method could translate to transformer architectures. 
Recent literature has demonstrated the efficacy of similar KD approaches in transformer architectures, suggesting that it would be possible to apply our framework to transformer architectures \citep{liu2024transkd, wang2022attention}.
However, we suggest that establishing if the distinctive intermediate-layer confidence trajectories that we see in CNN models (Figure \ref{fig:per_layer_conf_of_shortcuts}) is also mirrored in transformer architectures.

The requirement for a clean, curated dataset to train the teacher model presents a potential limitation, though our approach only necessitates a small amount of training data for the teacher network. 
While such an approach still imposes limitations and necessitates some degree of manual data curation and knowledge of possible sources of bias, the burden of doing so for this much smaller subset is significantly reduced compared to the full training dataset.

One interesting avenue for possible future work would be the use of generative models to create clean, synthetic training data for the teacher model. 
Additionally, self-supervised or unsupervised techniques for student training may provide a route to remove the need for an unbiased teacher model, addressing one of the primary limitations of this work. 

While our synthetic bias features provide a controlled experimental environment, a critical next step is the investigation of the effectiveness of our approach against a broader range of real-world medical image shortcuts, such as those related to patient demographics. 
This would further validate the practical utility of our method across diverse clinical contexts.

Our work advances both the theoretical understanding and practical mitigation of shortcut learning in medical image analysis. 
The demonstrated effectiveness of small specialist teachers and selective layer distillation provides a promising direction for developing robust medical AI systems that can generalize across healthcare environments. 
As these systems become increasingly prevalent in clinical settings, approaches like ours that can effectively prevent shortcut learning while maintaining high performance become crucial for ensuring safe and equitable healthcare delivery.


\acks{This work was supported by the UKRI EPSRC Centre for Doctoral Training in Applied Photonics [EP/S022821/1].}

%
\ethics{The work follows appropriate ethical standards in conducting research and writing the manuscript, following all applicable laws and regulations regarding treatment of animals or human subjects.}

\coi{We declare we don't have conflicts of interest.}

\data{
All datasets used in this study are publicly available at the following repositories:

CheXpert: \url{https://stanfordmlgroup.github.io/competitions/chexpert/}

ISIC: \url{https://challenge.isic-archive.com/data/\#2017}

SimBA: \url{https://borealisdata.ca/dataset.xhtml?persistentId=doi:10.5683/SP3/A9SOBZ}

MIMIC: \url{https://www.physionet.org/content/mimic-cxr-jpg/2.1.0/}

Fitzpatrick17k: \url{https://github.com/mattgroh/fitzpatrick17k}
\\
The class-balanced subsets used for training the teacher models can be reproduced following the methodology described in Section 3.
}

\bibliography{sample}


\clearpage
\appendix
\section{Shortcut reliance results in intermediate-layer overconfidence}

This section provides comprehensive layer-wise confidence analysis extending our main findings from Section \ref{sec:shortcut_ilc} to additional architectures and datasets, demonstrating the generalizability of our core observation that shortcut learning manifests distinctly across network layers across architectures and datasets. 

\begin{figure*}[ht!]
\includegraphics[width=\textwidth]{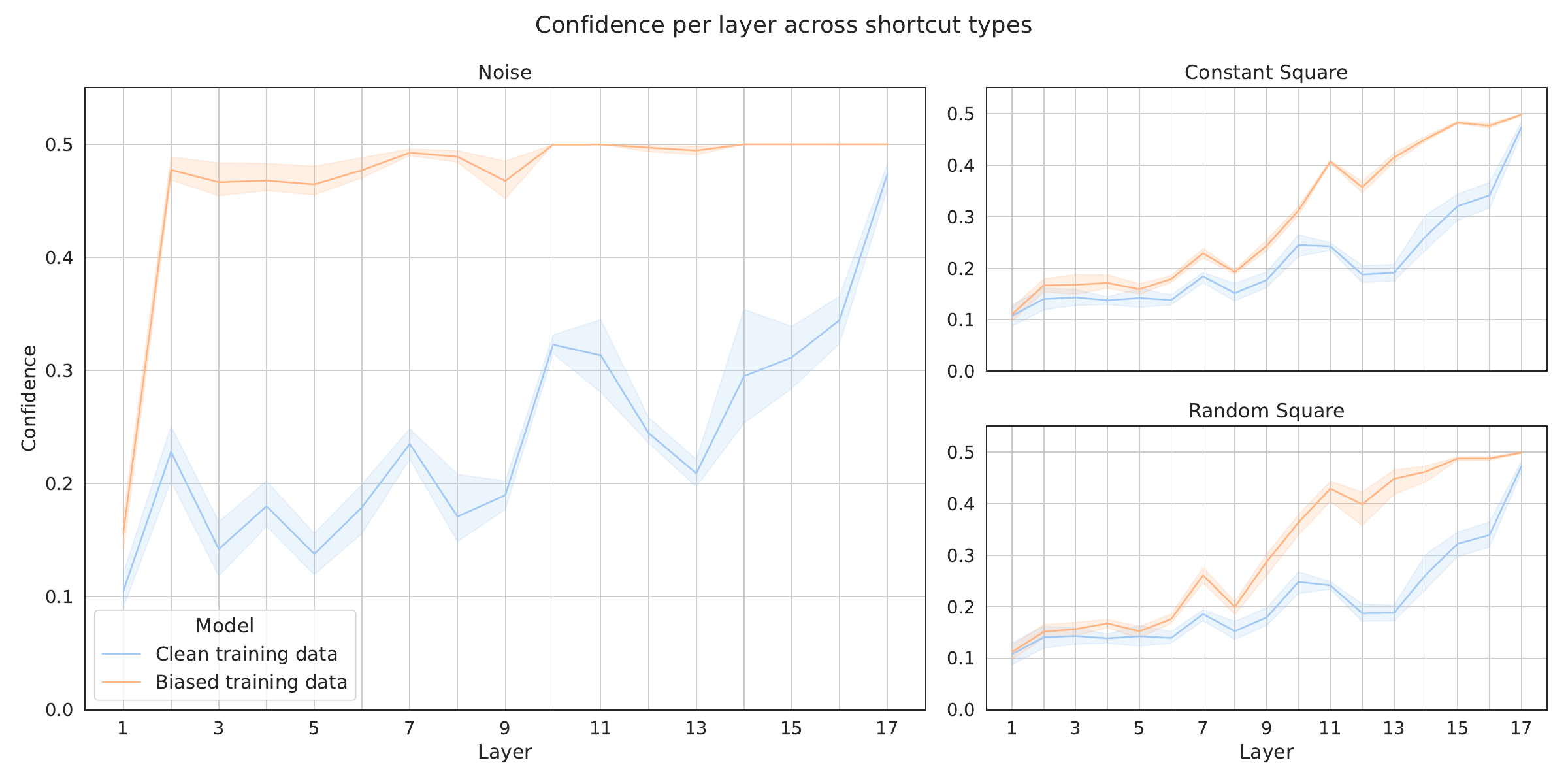}
\caption{Per-layer confidence of ResNet-18 students trained on ISIC data featuring various synthetic shortcuts.}\label{fig:isic-r18-per-layer-conf}
\end{figure*}

\begin{figure*}[th!]
\centering
\includegraphics[width=\textwidth]{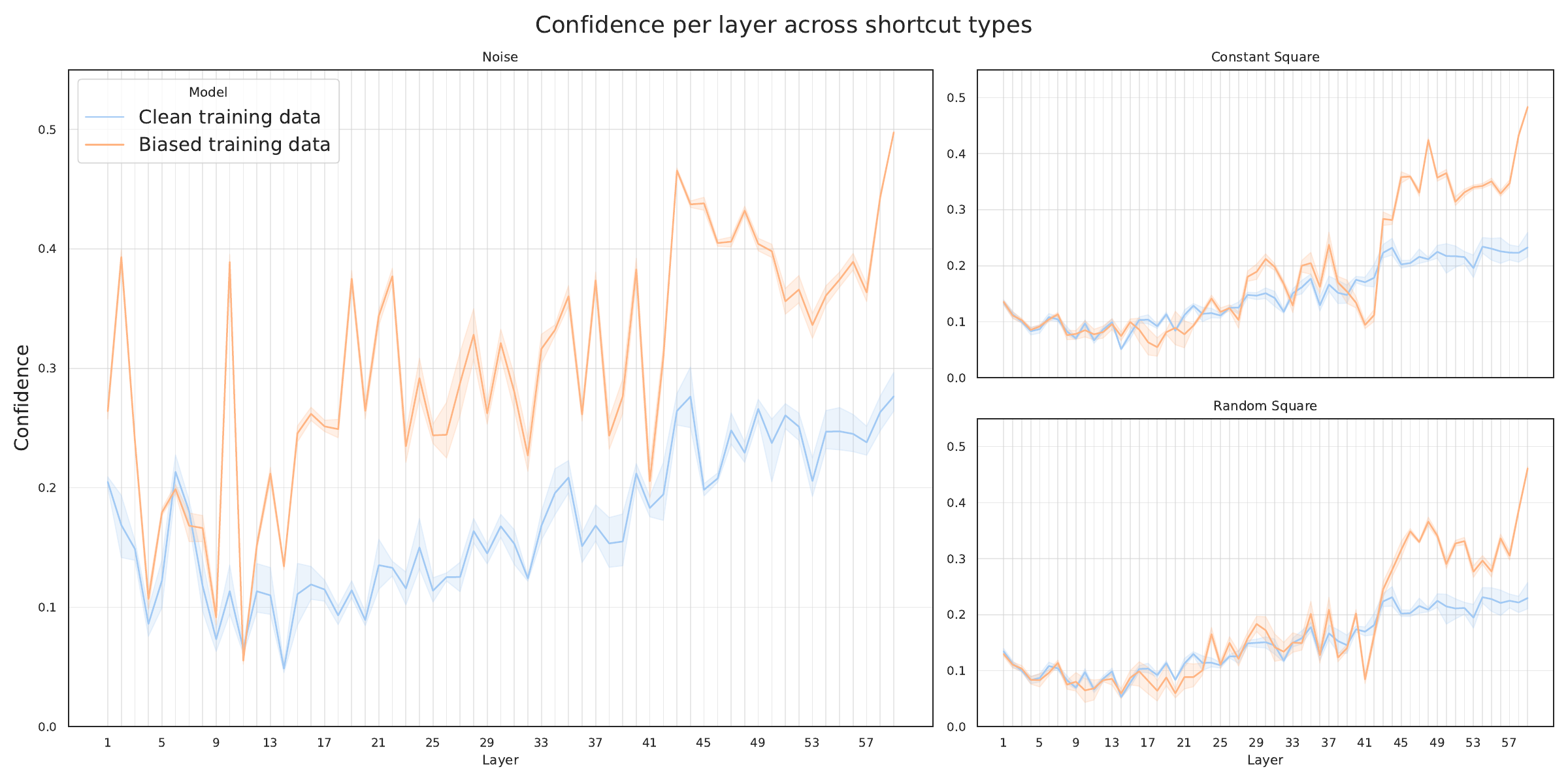}
\caption{Per-layer confidence of DenseNet-121 students trained on CheXpert data featuring various synthetic shortcuts. Shortcut prevalence in the train split is 100\%. In the validation and test split, the shortcut feature is balanced between classes.}\label{fig:chex-dn121-balanced-val-per-layer-conf}
\end{figure*}

Figures \ref{fig:isic-r18-per-layer-conf} and \ref{fig:chex-dn121-balanced-val-per-layer-conf} present intermediate layer confidence for ResNet-18 and DenseNet-121 architectures. In our main experiments, we use validation sets where shortcuts are balanced across classes (present equally in both positive and negative samples). Under these conditions, we observe that the overconfidence signal in DenseNet-121 is noticeably weaker than in ResNet-18. We hypothesize that the substantially larger capacity of DenseNet-121, combined with early stopping on validation data where shortcuts no longer correlate with class labels, prevents the network from fully developing shortcut dependencies.

\begin{figure*}[th!]
\centering
\includegraphics[width=\textwidth]{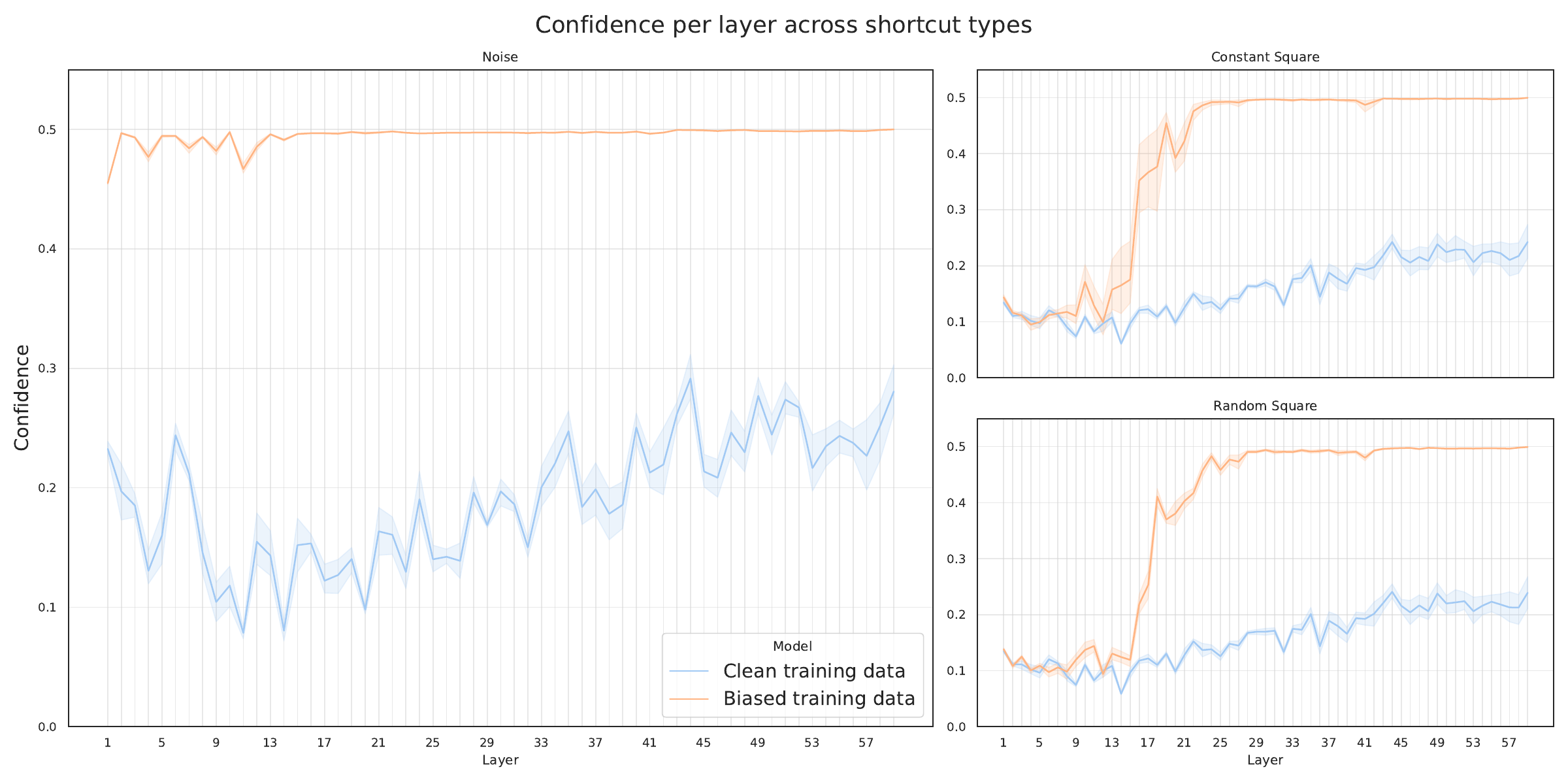}
\caption{Per-layer confidence of DenseNet-121 students trained on CheXpert data featuring various synthetic shortcuts. Shortcut prevalence in the train and validation splits is 100\%. In the test split, the shortcut feature is balanced between classes.}\label{fig:chexpert-dn121-per-layer-conf}
\end{figure*}

\begin{figure*}[th!]
\centering
\includegraphics[width=\textwidth]{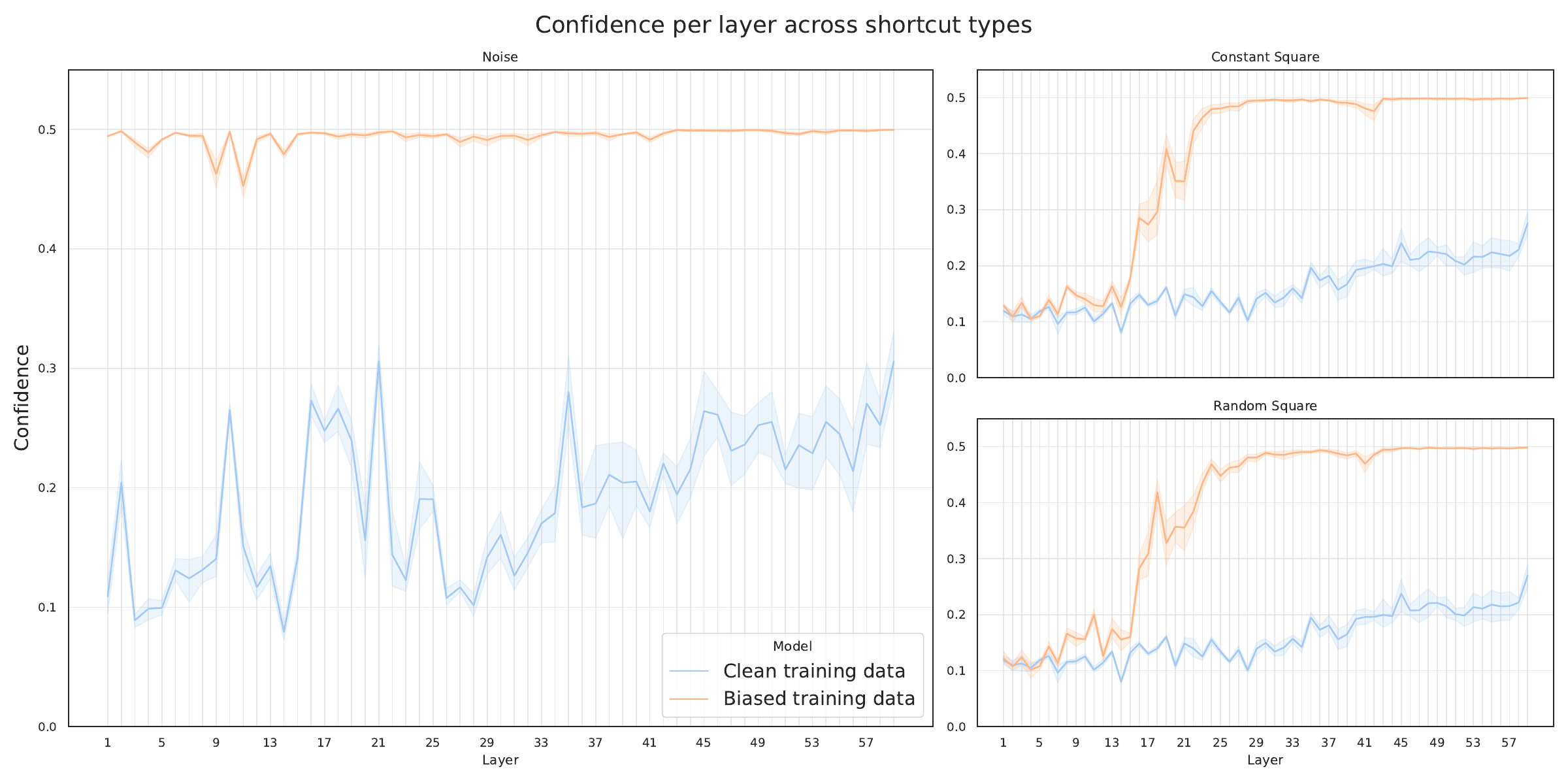}
\caption{Per-layer confidence of DenseNet-121 students trained on ISIC data featuring various synthetic shortcuts. Shortcut prevalence in the train and validation splits is 100\%. In the test split, the shortcut feature is balanced between classes.}\label{fig:isic-dn121-per-layer-conf}
\end{figure*}

To examine how validation set composition affects these patterns, Figures \ref{fig:chexpert-dn121-per-layer-conf} and \ref{fig:isic-dn121-per-layer-conf} show the same architectures trained with validation sets that maintain the same shortcut-class correlations as the training data. Under these conditions, the overconfidence signal becomes much more pronounced even in high-capacity models like DenseNet-121, as the validation setup no longer provides feedback that discourages shortcut reliance during training.


\section{Overall performance}%
This section provides detailed performance breakdowns, including AUC values and True Positive Rate (TPR) analysis for bias-aligned and bias-contrasting samples, supplementing the $\Delta$TPR analysis presented in Section \ref{sec:kd_main_res} of the main text.

Table \ref{tab:auc_table_new} presents AUC values across all experimental conditions, showing that our method consistently achieves performance competitive with the clean baseline even when trained on heavily biased data. Notably, our approach maintains stable performance across varying bias prevalence rates, while competing methods show significant degradation at higher bias levels.

Tables \ref{tab:chex_tpr_abs} and \ref{tab:isic_tpr_abs} provide absolute TPR values for CheXpert and ISIC datasets respectively, breaking down performance for bias-aligned samples (where shortcut presence matches training correlation with the class-label) and bias-contrasting samples (where shortcuts oppose training correlation with the class-label). We demonstrate that our method achieves more balanced performance across both sample types, indicating reduced reliance on shortcut features for prediction. We note that our method demonstrates lower TPR for bias-aligned samples compared to other methods, and that the better TPR seen in other models is likely a result of shortcut reliance, causing a TPR much higher than the clean baseline. We consistently observe that our method achieves a TPR for biased-aligned samples that is closest to the clean baseline.

\begin{table*}[!ht]
\centering
\caption{$AUC\uparrow$ of a ResNet-18 trained on data with various bias prevalence rates. Results are presented as Mean$\pm$Std over 5-fold cross-validation. Models are marked as \textbf{best} and \underline{second-best}. 
}
\footnotesize
\begin{tabular}{c c c c c c c c}
\midrule
Prev. & \multirow{2}{*}{Model} & \multicolumn{3}{c}{CheXpert} &\multicolumn{3}{c}{ISIC} \\ \cmidrule(lr){3-5} \cmidrule{6-8}
 (\%) & & Noise & Square (C) & Square (R) & Noise & Square (C) & Square (R) \\
 \midrule
 0 & Baseline & 0.709\scriptsize$\pm$0.024 & 0.755\scriptsize$\pm$0.013 & 0.752\scriptsize$\pm$0.013 & 0.749\scriptsize$\pm$0.024 & 0.809\scriptsize$\pm$0.019  & 0.808\scriptsize$\pm$0.019  \\
 
\midrule

\multirow{6}{*}{100}
 & ERM  & 0.489\scriptsize$\pm$0.012 & 0.533\scriptsize$\pm$0.007 & 0.554\scriptsize$\pm$0.005 & \underline{0.521\scriptsize$\pm$0.010} & 0.600\scriptsize$\pm$0.011 &  0.612\scriptsize$\pm$0.007 \\
 & MixUp & 0.509\scriptsize$\pm$0.007 & 0.539\scriptsize$\pm$0.008 & 0.549\scriptsize$\pm$0.005 & 0.489\scriptsize$\pm$0.012 & 0.555\scriptsize$\pm$0.028 & 0.591\scriptsize$\pm$0.014 \\
 & CutOut  & 0.498\scriptsize$\pm$0.029 & 0.548\scriptsize$\pm$0.010 & 0.583\scriptsize$\pm$0.011 & 0.521\scriptsize$\pm$0.023 & 0.627\scriptsize$\pm$0.013 & 0.639\scriptsize$\pm$0.014 \\
 & CutMix & \underline{0.529\scriptsize$\pm$0.007}  & \underline{0.680\scriptsize$\pm$0.025} & \underline{0.759\scriptsize$\pm$0.011} & 0.516\scriptsize$\pm$0.015 & \underline{0.753\scriptsize$\pm$0.038} & \underline{0.784\scriptsize$\pm$0.017}  \\
 & Aug & 0.483\scriptsize$\pm$0.009 &  0.585\scriptsize$\pm$0.017 & 0.555\scriptsize$\pm$0.010 & 0.518\scriptsize$\pm$0.006 & 0.731\scriptsize$\pm$0.017 & 0.626\scriptsize$\pm$0.014 \\
 & Ours & \textbf{0.689\scriptsize$\pm$0.044} & \textbf{0.747\scriptsize$\pm$0.008} & \textbf{0.761\scriptsize$\pm$0.010} & \textbf{0.775\scriptsize$\pm$0.023} & \textbf{0.777\scriptsize$\pm$0.024} & \textbf{0.806\scriptsize$\pm$0.016} \\
 
\midrule

\multirow{8}{*}{95}
 & ERM & 0.579\scriptsize$\pm$0.009 & 0.597\scriptsize$\pm$0.007 & 0.595\scriptsize$\pm$0.015 & 0.602\scriptsize$\pm$0.032 & 0.645\scriptsize$\pm$0.024 & 0.651\scriptsize$\pm$0.028  \\
 & MixUp & 0.559\scriptsize$\pm$0.008 & 0.578\scriptsize$\pm$0.008 & 0.603\scriptsize$\pm$0.024 & 0.542\scriptsize$\pm$0.043  & 0.609\scriptsize$\pm$0.021 & 0.603\scriptsize$\pm$0.024  \\
 & CutOut & 0.568\scriptsize$\pm$0.013 & 0.599\scriptsize$\pm$0.013 & 0.611\scriptsize$\pm$0.040 & \underline{0.608\scriptsize$\pm$0.024} & 0.641\scriptsize$\pm$0.011  &  0.611\scriptsize$\pm$0.040 \\
 & CutMix &  0.567\scriptsize$\pm$0.020 & \underline{0.739\scriptsize$\pm$0.024} & \textbf{0.760\scriptsize$\pm$0.020} & 0.587\scriptsize$\pm$0.029 &  \underline{0.768\scriptsize$\pm$0.031} & \underline{0.760\scriptsize$\pm$0.020}  \\
 & Aug & \underline{0.578\scriptsize$\pm$0.014} & 0.628\scriptsize$\pm$0.028 & 0.597\scriptsize$\pm$0.015 & 0.608\scriptsize$\pm$0.026 &  0.745\scriptsize$\pm$0.010 & 0.633\scriptsize$\pm$0.013  \\
 & GDRO & 0.566\scriptsize$\pm$0.008 & 0.585\scriptsize$\pm$0.010 & 0.613\scriptsize$\pm$0.014 & 0.595\scriptsize$\pm$0.012 & 0.631\scriptsize$\pm$0.008 & 0.673\scriptsize$\pm$0.015  \\
 & JTT & 0.565\scriptsize$\pm$0.007 & 0.588\scriptsize$\pm$0.004 & 0.602\scriptsize$\pm$0.018 & 0.591\scriptsize$\pm$0.026 & 0.630\scriptsize$\pm$0.015& 0.672\scriptsize$\pm$0.016 \\
 & Ours & \textbf{0.693\scriptsize$\pm$0.029} & \textbf{0.756\scriptsize$\pm$0.005} & \underline{0.756\scriptsize$\pm$0.011} & \textbf{0.778\scriptsize$\pm$0.020} & \textbf{0.797\scriptsize$\pm$0.015} & \textbf{0.803\scriptsize$\pm$0.009} \\

\midrule

\multirow{8}{*}{85}
 & ERM & 0.621\scriptsize$\pm$0.018 & 0.625\scriptsize$\pm$0.018 & 0.661\scriptsize$\pm$0.013 & \underline{0.694\scriptsize$\pm$0.018} & 0.751\scriptsize$\pm$0.016 & 0.748\scriptsize$\pm$0.030 \\
 & MixUp & 0.598\scriptsize$\pm$0.017 & 0.630\scriptsize$\pm$0.030 & 0.697\scriptsize$\pm$0.021 & 0.664\scriptsize$\pm$0.034 & 0.687\scriptsize$\pm$0.019 & 0.706\scriptsize$\pm$0.052 \\
 & CutOut& 0.596\scriptsize$\pm$0.018 & 0.639\scriptsize$\pm$0.026 &0.683\scriptsize$\pm$0.018& 0.663\scriptsize$\pm$0.035 & 0.731\scriptsize$\pm$0.040 & 0.739\scriptsize$\pm$0.029 \\
 & CutMix & 0.604\scriptsize$\pm$0.015 & \underline{0.742\scriptsize$\pm$0.049} & \textbf{0.766\scriptsize$\pm$0.016} & 0.634\scriptsize$\pm$0.015 & 0.766\scriptsize$\pm$0.039 & \textbf{0.796\scriptsize$\pm$0.019} \\
 & Aug & 0.636\scriptsize$\pm$0.034 & 0.677\scriptsize$\pm$0.040 & 0.661\scriptsize$\pm$0.035 & 0.667\scriptsize$\pm$0.039 & \underline{0.768\scriptsize$\pm$0.018} & 0.694\scriptsize$\pm$0.041\\
 & GDRO & 0.616\scriptsize$\pm$0.012 & 0.649\scriptsize$\pm$0.015 & 0.725\scriptsize$\pm$0.015 & 0.655\scriptsize$\pm$0.027 & 0.732\scriptsize$\pm$0.022 & 0.780\scriptsize$\pm$0.018  \\
 & JTT &\textbf{0.780\scriptsize$\pm$0.018}  & 0.659\scriptsize$\pm$0.008 & 0.699\scriptsize$\pm$0.010 &  0.656\scriptsize$\pm$0.025& 0.721\scriptsize$\pm$0.022&  0.771\scriptsize$\pm$0.018 \\
 & Ours & \underline{0.708\scriptsize$\pm$0.037} &  \textbf{0.756\scriptsize$\pm$0.008} & \underline{0.760\scriptsize$\pm$0.015} & \textbf{0.798\scriptsize$\pm$0.024} & \textbf{0.782\scriptsize$\pm$0.031} & \underline{0.787\scriptsize$\pm$0.026}  \\
 
\midrule

\multirow{8}{*}{75}
 & ERM & 0.681\scriptsize$\pm$0.023 & 0.686\scriptsize$\pm$0.038 & 0.718\scriptsize$\pm$0.015 & 0.711\scriptsize$\pm$0.028 & 0.763\scriptsize$\pm$0.013 & 0.784\scriptsize$\pm$0.013 \\
 & MixUp & 0.677\scriptsize$\pm$0.025 & 0.705\scriptsize$\pm$0.030  & 0.746\scriptsize$\pm$0.026 & 0.673\scriptsize$\pm$0.018 & 0.720\scriptsize$\pm$0.021 & 0.757\scriptsize$\pm$0.016 \\
 & CutOut & 0.675\scriptsize$\pm$0.027 & 0.707\scriptsize$\pm$0.026 & 0.744\scriptsize$\pm$0.011 & 0.715\scriptsize$\pm$0.034 & 0.726\scriptsize$\pm$0.020 & 0.757\scriptsize$\pm$0.026 \\
 & CutMix & 0.662\scriptsize$\pm$0.029 & \underline{0.748\scriptsize$\pm$0.027} & 0.759\scriptsize$\pm$0.031 & 0.696\scriptsize$\pm$0.025 & 0.768\scriptsize$\pm$0.030 & 0.781\scriptsize$\pm$0.029 \\
 
 & Aug & 0.676\scriptsize$\pm$0.026 & 0.712\scriptsize$\pm$0.041 & 0.701\scriptsize$\pm$0.041 & \underline{0.722\scriptsize$\pm$0.029} & 0.766\scriptsize$\pm$0.022 & 0.731\scriptsize$\pm$0.035 \\
 & GDRO & \underline{0.692\scriptsize$\pm$0.010} & 0.731\scriptsize$\pm$0.012 & \textbf{0.768\scriptsize$\pm$0.010} & 0.696\scriptsize$\pm$0.013 & \underline{0.770\scriptsize$\pm$0.008} & \textbf{0.796\scriptsize$\pm$0.008} \\
 & JTT & 0.687\scriptsize$\pm$0.020 & 0.730\scriptsize$\pm$0.017 & 0.751\scriptsize$\pm$0.026 & 0.694\scriptsize$\pm$0.009 & 0.769\scriptsize$\pm$0.011 & \underline{0.794\scriptsize$\pm$0.012} \\
 & Ours & \textbf{0.709\scriptsize$\pm$0.039} & \textbf{0.760\scriptsize$\pm$0.015} & \underline{0.762\scriptsize$\pm$0.017} & \textbf{0.792\scriptsize$\pm$0.023} & \textbf{0.773\scriptsize$\pm$0.013} & 0.781\scriptsize$\pm$0.029 \\
 \bottomrule
\\
\end{tabular}

\label{tab:auc_table_new}
\end{table*}

\begin{table*}[!ht]
\centering
\caption{$TPR\uparrow$ of bias-aligned and bias-contrasting samples for a ResNet-18 trained on CheXpert data with various bias prevalence rates. Results are presented as Mean$\pm$Std over 5-fold cross-validation. Models are marked as \textbf{best} and \underline{second-best}. When multiple models achieve identical performance, all are highlighted. 
}
\footnotesize
\begin{tabular}{c c c c c c c c}
\midrule
Prev. & \multirow{2}{*}{Model} & \multicolumn{2}{c}{Noise} & \multicolumn{2}{c}{Square (C)} & \multicolumn{2}{c}{Square (R)}\\ \cmidrule(lr){3-4} \cmidrule(lr){5-6}  \cmidrule{7-8}
 (\%) & & Bias-Aligned & Bias-Contrasting & Bias-Aligned & Bias-Contrasting & Bias-Aligned & Bias-Contrasting \\
 \midrule
 0 & Baseline & 0.942\scriptsize$\pm$0.036 & 0.811\scriptsize$\pm$0.069 & 0.804\scriptsize$\pm$0.080 & 0.811\scriptsize$\pm$0.069 & 0.822\scriptsize$\pm$0.069 & 0.811\scriptsize$\pm$0.069  \\
 
\midrule

\multirow{6}{*}{100}

 & ERM  & \textbf{1.000\scriptsize$\pm$0.000} & 0.000\scriptsize$\pm$0.000 & \textbf{1.000\scriptsize$\pm$0.000} & 0.000\scriptsize$\pm$0.000  & \textbf{0.995\scriptsize$\pm$0.005} & 0.004\scriptsize$\pm$0.007 \\
 & MixUp & \textbf{1.000\scriptsize$\pm$0.000} & \underline{0.013\scriptsize$\pm$0.026} & \underline{0.998\scriptsize$\pm$0.003} & 0.000\scriptsize$\pm$0.000 & 0.974\scriptsize$\pm$0.018 & 0.004\scriptsize$\pm$0.007 \\
 & CutOut  & \textbf{1.000\scriptsize$\pm$0.000} & 0.001\scriptsize$\pm$0.001 & \textbf{1.000\scriptsize$\pm$0.000} & 0.000\scriptsize$\pm$0.000 & \underline{0.982\scriptsize$\pm$0.011} & 0.004\scriptsize$\pm$0.009 \\
 & CutMix  & \textbf{1.000\scriptsize$\pm$0.000} & 0.007\scriptsize$\pm$0.005 & 0.929\scriptsize$\pm$0.045 & \underline{0.427\scriptsize$\pm$0.114} & 0.894\scriptsize$\pm$0.014 & \underline{0.765\scriptsize$\pm$0.015} \\
 & Aug & \underline{0.957\scriptsize$\pm$0.092} & 0.000\scriptsize$\pm$0.000 & 0.996\scriptsize$\pm$0.004 & 0.017\scriptsize$\pm$0.015 & \underline{0.982\scriptsize$\pm$0.011} & 0.004\scriptsize$\pm$0.009  \\
 & Ours & 0.942\scriptsize$\pm$0.033 & \textbf{0.565\scriptsize$\pm$0.161} & 0.862\scriptsize$\pm$0.005 & \textbf{0.783\scriptsize$\pm$0.017} & 0.839\scriptsize$\pm$0.063 & \textbf{0.805\scriptsize$\pm$0.050} \\
 
\midrule

\multirow{8}{*}{95}
 & ERM  & 0.995\scriptsize$\pm$0.010 & 0.055\scriptsize$\pm$0.022 & 0.993\scriptsize$\pm$0.011 & 0.081\scriptsize$\pm$0.088 & 0.973\scriptsize$\pm$0.014 & 0.182\scriptsize$\pm$0.105 \\
 & MixUp & 0.997\scriptsize$\pm$0.006 & 0.070\scriptsize$\pm$0.107 & 0.992\scriptsize$\pm$0.007 & 0.005\scriptsize$\pm$0.010 & 0.948\scriptsize$\pm$0.015 & 0.253\scriptsize$\pm$0.100 \\
 & CutOut  & 0.997\scriptsize$\pm$0.004 & 0.047\scriptsize$\pm$0.051 & 0.987\scriptsize$\pm$0.013 & 0.074\scriptsize$\pm$0.061 & 0.959\scriptsize$\pm$0.007 & 0.330\scriptsize$\pm$0.127 \\
 & CutMix  & \underline{0.999\scriptsize$\pm$0.001} & 0.025\scriptsize$\pm$0.029 & 0.908\scriptsize$\pm$0.022 & \underline{0.583\scriptsize$\pm$0.047} & 0.845\scriptsize$\pm$0.029 & \underline{0.709\scriptsize$\pm$0.070} \\
 & Aug & 0.997\scriptsize$\pm$0.003  & \underline{0.098\scriptsize$\pm$0.075} & 0.962\scriptsize$\pm$0.039 & 0.183\scriptsize$\pm$0.132 & \underline{0.978\scriptsize$\pm$0.014} & 0.159\scriptsize$\pm$0.079  \\
 & JTT& \textbf{1.000\scriptsize$\pm$0.000} & 0.018\scriptsize$\pm$0.011 & \textbf{0.999\scriptsize$\pm$0.001} & 0.054\scriptsize$\pm$0.031 & 0.956\scriptsize$\pm$0.080 & 0.283\scriptsize$\pm$0.041 \\
 & GDRO& \textbf{1.000\scriptsize$\pm$0.000} & 0.014\scriptsize$\pm$0.010 & \underline{0.997\scriptsize$\pm$0.002} & 0.019\scriptsize$\pm$0.014 & \textbf{0.981\scriptsize$\pm$0.011} & 0.279\scriptsize$\pm$0.096 \\
 & Ours & 0.935\scriptsize$\pm$0.037 & \textbf{0.563\scriptsize$\pm$0.092} & 0.862\scriptsize$\pm$0.031 & \textbf{0.773\scriptsize$\pm$0.019} & 0.810\scriptsize$\pm$0.051 & \textbf{0.775\scriptsize$\pm$0.021} \\

\midrule

\multirow{8}{*}{85}
 & ERM  & 0.960\scriptsize$\pm$0.039 & 0.215\scriptsize$\pm$0.167 & 0.974\scriptsize$\pm$0.013 & 0.239\scriptsize$\pm$0.115 & 0.926\scriptsize$\pm$0.047 &  0.503\scriptsize$\pm$0.106 \\
 & MixUp & 0.985\scriptsize$\pm$0.011 & 0.286\scriptsize$\pm$0.119 & 0.962\scriptsize$\pm$0.035 & 0.284\scriptsize$\pm$0.117 & 0.903\scriptsize$\pm$0.022 & 0.566\scriptsize$\pm$0.031 \\
 & CutOut  & \textbf{0.988\scriptsize$\pm$0.009} & 0.178\scriptsize$\pm$0.072 & \underline{0.977\scriptsize$\pm$0.014} & 0.320\scriptsize$\pm$0.204 & \textbf{0.938\scriptsize$\pm$0.029} & 0.557\scriptsize$\pm$0.127 \\
 & CutMix  & 0.979\scriptsize$\pm$0.021 & 0.246\scriptsize$\pm$0.106 & 0.886\scriptsize$\pm$0.039 & \underline{0.675\scriptsize$\pm$0.050} & 0.835\scriptsize$\pm$0.036 & \underline{0.737\scriptsize$\pm$0.037} \\
 & Aug & 0.957\scriptsize$\pm$0.039 & \underline{0.292\scriptsize$\pm$0.208} & 0.954\scriptsize$\pm$0.007 & 0.437\scriptsize$\pm$0.197 & 0.927\scriptsize$\pm$0.030 & 0.397\scriptsize$\pm$0.144  \\
 & JTT& \underline{0.987\scriptsize$\pm$0.005} & 0.268\scriptsize$\pm$0.034 & 0.968\scriptsize$\pm$0.008 & 0.435\scriptsize$\pm$0.030 & \underline{0.933\scriptsize$\pm$0.029} & 0.555\scriptsize$\pm$0.144 \\
 & GDRO& 0.983\scriptsize$\pm$0.006 & 0.224\scriptsize$\pm$0.033 &\textbf{0.977\scriptsize$\pm$0.011}  & 0.353\scriptsize$\pm$0.057 & 0.922\scriptsize$\pm$0.012 & 0.654\scriptsize$\pm$0.071 \\
 & Ours & 0.929\scriptsize$\pm$0.042 & \textbf{0.581\scriptsize$\pm$0.119} & 0.861\scriptsize$\pm$0.060 & \textbf{0.755\scriptsize$\pm$0.030} & 0.860\scriptsize$\pm$0.026 & \textbf{0.801\scriptsize$\pm$0.044} \\
 
\midrule

\multirow{8}{*}{75}
 & ERM  & 0.879\scriptsize$\pm$0.099 & 0.434\scriptsize$\pm$0.193 & 0.903\scriptsize$\pm$0.125 & 0.516\scriptsize$\pm$0.077 & \textbf{0.910\scriptsize$\pm$0.025} & 0.708\scriptsize$\pm$0.078 \\
 & MixUp & 0.944\scriptsize$\pm$0.017 & \textbf{0.564\scriptsize$\pm$0.130} & 0.888\scriptsize$\pm$0.066 & 0.557\scriptsize$\pm$0.088 & 0.849\scriptsize$\pm$0.044 & 0.692\scriptsize$\pm$0.032 \\
 & CutOut  & 0.941\scriptsize$\pm$0.043 & 0.481\scriptsize$\pm$0.087 & 0.855\scriptsize$\pm$0.119 & 0.494\scriptsize$\pm$0.187 & \underline{0.897\scriptsize$\pm$0.021} & 0.729\scriptsize$\pm$0.063  \\
 & CutMix  & 0.926\scriptsize$\pm$0.045 & 0.399\scriptsize$\pm$0.085 & 0.852\scriptsize$\pm$0.072 & \underline{0.714\scriptsize$\pm$0.091} & 0.819\scriptsize$\pm$0.028 & \underline{0.763\scriptsize$\pm$0.012} \\
 & Aug & 0.909\scriptsize$\pm$0.051 & 0.463\scriptsize$\pm$0.171 & 0.878\scriptsize$\pm$0.059 & 0.470\scriptsize$\pm$0.205 & 0.849\scriptsize$\pm$0.032 & 0.505\scriptsize$\pm$0.071  \\
 & JTT& \underline{0.949\scriptsize$\pm$0.023} & 0.497\scriptsize$\pm$0.052 & \textbf{0.928\scriptsize$\pm$0.025} & 0.667\scriptsize$\pm$0.036 & 0.871\scriptsize$\pm$0.028 & 0.725\scriptsize$\pm$0.060 \\
 & GDRO& \textbf{0.951\scriptsize$\pm$0.004} & 0.444\scriptsize$\pm$0.015 & \underline{0.904\scriptsize$\pm$0.023} & 0.575\scriptsize$\pm$0.022 & 0.866\scriptsize$\pm$0.022 & 0.752\scriptsize$\pm$0.019  \\
 & Ours & 0.926\scriptsize$\pm$0.040 & \underline{0.561\scriptsize$\pm$0.162} & 0.878\scriptsize$\pm$0.029 & \textbf{0.751\scriptsize$\pm$0.042} & 0.854\scriptsize$\pm$0.044 & \textbf{0.793\scriptsize$\pm$0.056} \\
 \bottomrule
\end{tabular}

\label{tab:chex_tpr_abs}
\end{table*}

\begin{table*}[!ht]
\centering
\caption{$TPR\uparrow$ of bias-aligned and bias-contrasting samples for a ResNet-18 trained on ISIC data with various bias prevalence rates. Results are presented as Mean$\pm$Std over 5-fold cross-validation. Models are marked as \textbf{best} and \underline{second-best}. When multiple models achieve identical performance, all are highlighted. 
}
\footnotesize
\begin{tabular}{c c c c c c c c}
\midrule
Prev. & \multirow{2}{*}{Model} & \multicolumn{2}{c}{Noise} & \multicolumn{2}{c}{Square (C)} & \multicolumn{2}{c}{Square (R)}\\ \cmidrule(lr){3-4} \cmidrule(lr){5-6}  \cmidrule{7-8}
 (\%) & & Bias-Aligned & Bias-Contrasting & Bias-Aligned & Bias-Contrasting & Bias-Aligned & Bias-Contrasting \\
 \midrule
 0 & Baseline & 0.449\scriptsize$\pm$0.113 & 0.858\scriptsize$\pm$0.027 & 0.802\scriptsize$\pm$0.030 & 0.858\scriptsize$\pm$0.027 & 0.808\scriptsize$\pm$0.036 &  0.858\scriptsize$\pm$0.027 \\
 
\midrule

\multirow{6}{*}{100}
 & ERM  & \textbf{1.000\scriptsize$\pm$0.000} & \underline{0.000\scriptsize$\pm$0.000} &  \textbf{1.000\scriptsize$\pm$0.000} & 0.223\scriptsize$\pm$0.093 & \textbf{1.000\scriptsize$\pm$0.000} & 0.156\scriptsize$\pm$0.168 \\
 & MixUp & \underline{0.998\scriptsize$\pm$0.004} & \underline{0.000\scriptsize$\pm$0.000} & \textbf{1.000\scriptsize$\pm$0.000} & 0.127\scriptsize$\pm$0.150 & \underline{0.990\scriptsize$\pm$0.012} & 0.238\scriptsize$\pm$0.146 \\
 & CutOut  & \textbf{1.000\scriptsize$\pm$0.000} & \underline{0.000\scriptsize$\pm$0.000} & \textbf{1.000\scriptsize$\pm$0.000} & 0.552\scriptsize$\pm$0.065 & 0.986\scriptsize$\pm$0.016 & 0.712\scriptsize$\pm$0.066 \\
 & CutMix  & \textbf{1.000\scriptsize$\pm$0.000} & \underline{0.000\scriptsize$\pm$0.000} & 0.905\scriptsize$\pm$0.032 & 0.546\scriptsize$\pm$0.083 & 0.856\scriptsize$\pm$0.021 &\underline{0.729\scriptsize$\pm$0.050}  \\
 & Aug  & \textbf{1.000\scriptsize$\pm$0.000} & \underline{0.000\scriptsize$\pm$0.000} & \underline{0.961\scriptsize$\pm$0.019} & \underline{0.800\scriptsize$\pm$0.062} & 0.971\scriptsize$\pm$0.023 &  0.467\scriptsize$\pm$0.136 \\
 & Ours & 0.792\scriptsize$\pm$0.066 & \textbf{0.771\scriptsize$\pm$0.09} & 0.829\scriptsize$\pm$0.044 & \textbf{0.852\scriptsize$\pm$0.028} & 0.829\scriptsize$\pm$0.042 & \textbf{0.900\scriptsize$\pm$0.016} \\
 
\midrule

\multirow{8}{*}{95}
 & ERM  & \textbf{1.000\scriptsize$\pm$0.000} & 0.041\scriptsize$\pm$0.019 & \underline{0.993\scriptsize$\pm$0.010} & 0.132\scriptsize$\pm$0.026 & \underline{0.987\scriptsize$\pm$0.020} & 0.285\scriptsize$\pm$0.097 \\
 & MixUp & 0.973\scriptsize$\pm$0.047 & 0.021\scriptsize$\pm$0.019 &  \textbf{0.998\scriptsize$\pm$0.004} & 0.062\scriptsize$\pm$0.035 & 0.984\scriptsize$\pm$0.020 & 0.229\scriptsize$\pm$0.081 \\
 & CutOut  & \textbf{1.000\scriptsize$\pm$0.000} & 0.054\scriptsize$\pm$0.047 & 0.985\scriptsize$\pm$0.023 & 0.406\scriptsize$\pm$0.263 & 0.973\scriptsize$\pm$0.027 &  0.544\scriptsize$\pm$0.144\\
 & CutMix  & \underline{0.998\scriptsize$\pm$0.004} & \underline{0.076\scriptsize$\pm$0.079} & 0.909\scriptsize$\pm$0.017 & 0.598\scriptsize$\pm$0.098 &0.880\scriptsize$\pm$0.024  & \underline{0.757\scriptsize$\pm$0.050} \\
 & Aug & 0.996\scriptsize$\pm$0.008 & 0.062\scriptsize$\pm$0.059 & 0.965\scriptsize$\pm$0.015 & \underline{0.761\scriptsize$\pm$0.079} & 0.985\scriptsize$\pm$0.014 &  0.429\scriptsize$\pm$0.126 \\
 & GDRO  & 0.996\scriptsize$\pm$0.005 & 0.029\scriptsize$\pm$0.017 & 0.987\scriptsize$\pm$0.005 & 0.188\scriptsize$\pm$0.035 & 0.980\scriptsize$\pm$0.010 & 0.503\scriptsize$\pm$0.024 \\
 & JTT  & \textbf{1.000\scriptsize$\pm$0.000} & 0.054\scriptsize$\pm$0.031 & 0.991\scriptsize$\pm$0.011 & 0.198\scriptsize$\pm$0.049 & \textbf{0.987\scriptsize$\pm$0.014} &  0.482\scriptsize$\pm$0.070\\
 & Ours & 0.858\scriptsize$\pm$0.042  & \textbf{0.781\scriptsize$\pm$0.062} & 0.898\scriptsize$\pm$0.024  & \textbf{0.798\scriptsize$\pm$0.035} & 0.873\scriptsize$\pm$0.031 & \textbf{0.821\scriptsize$\pm$0.036} \\

\midrule

\multirow{8}{*}{85}
 & ERM  & 0.958\scriptsize$\pm$0.008 & \underline{0.685\scriptsize$\pm$0.034} & 0.931\scriptsize$\pm$0.020 & 0.555\scriptsize$\pm$0.104 & \textbf{0.954\scriptsize$\pm$0.011} & 0.660\scriptsize$\pm$0.083 \\
 & MixUp & 0.952\scriptsize$\pm$0.016 & 0.398\scriptsize$\pm$0.159 & 0.939\scriptsize$\pm$0.031 & 0.449\scriptsize$\pm$0.100 & 0.905\scriptsize$\pm$0.043 & 0.511\scriptsize$\pm$0.084 \\
 & CutOut  & 0.952\scriptsize$\pm$0.024 & 0.442\scriptsize$\pm$0.215 & \textbf{0.954\scriptsize$\pm$0.022} & 0.483\scriptsize$\pm$0.155 & \underline{0.933\scriptsize$\pm$0.015} & 0.683\scriptsize$\pm$0.130 \\
 & CutMix  & 0.970\scriptsize$\pm$0.025 & 0.317\scriptsize$\pm$0.113 & 0.895\scriptsize$\pm$0.033 & 0.681\scriptsize$\pm$0.074 & 0.871\scriptsize$\pm$0.025 & 0.785\scriptsize$\pm$0.040 \\
 & Aug &0.950\scriptsize$\pm$0.029  & 0.274\scriptsize$\pm$0.185 & 0.883\scriptsize$\pm$0.056 & \underline{0.768\scriptsize$\pm$0.066} & 0.913\scriptsize$\pm$0.051 &  0.594\scriptsize$\pm$0.123 \\
 & GDRO  & \textbf{0.980\scriptsize$\pm$0.010} & 0.283\scriptsize$\pm$0.083 & 0.945\scriptsize$\pm$0.034 & 0.609\scriptsize$\pm$0.052 & 0.903\scriptsize$\pm$0.039 & \underline{0.806\scriptsize$\pm$0.017}\\
 & JTT  & \underline{0.978\scriptsize$\pm$0.011} & 0.275\scriptsize$\pm$0.119 & \underline{0.950\scriptsize$\pm$0.019 }& 0.600\scriptsize$\pm$0.058 & 0.911\scriptsize$\pm$0.029 & 0.753\scriptsize$\pm$0.037 \\
 & Ours & 0.855\scriptsize$\pm$0.033 & \textbf{0.792\scriptsize$\pm$0.100} & 0.863\scriptsize$\pm$0.047 & \textbf{0.817\scriptsize$\pm$0.032} & 0.861\scriptsize$\pm$0.031 & \textbf{0.811\scriptsize$\pm$0.086} \\
 
\midrule

\multirow{8}{*}{75}
 & ERM  & 0.923\scriptsize$\pm$0.022 & \underline{0.670\scriptsize$\pm$0.048} & 0.829\scriptsize$\pm$0.041 & 0.616\scriptsize$\pm$0.051 & 0.894\scriptsize$\pm$0.019 & 0.721\scriptsize$\pm$0.083 \\
 & MixUp & \textbf{0.948\scriptsize$\pm$0.019} & 0.477\scriptsize$\pm$0.168 & \underline{0.927\scriptsize$\pm$0.035} & 0.645\scriptsize$\pm$0.126 & \textbf{0.902\scriptsize$\pm$0.017} & 0.732\scriptsize$\pm$0.040 \\
 & CutOut  & 0.933\scriptsize$\pm$0.012 & 0.668\scriptsize$\pm$0.189 & \textbf{0.956\scriptsize$\pm$0.014} & 0.816\scriptsize$\pm$0.047 & \underline{0.896\scriptsize$\pm$0.041} & 0.724\scriptsize$\pm$0.136 \\
 & CutMix  &  0.910\scriptsize$\pm$0.040 & 0.580\scriptsize$\pm$0.098 & 0.867\scriptsize$\pm$0.029 & 0.714\scriptsize$\pm$0.056 & 0.863\scriptsize$\pm$0.026 & 0.793\scriptsize$\pm$0.056 \\
 & Aug & 0.915\scriptsize$\pm$0.034 & 0.591\scriptsize$\pm$0.155 & 0.863\scriptsize$\pm$0.047 &  0.714\scriptsize$\pm$0.078 & 0.894\scriptsize$\pm$0.048 &  0.636\scriptsize$\pm$0.118 \\
 & GDRO  & \underline{0.942\scriptsize$\pm$0.019} & 0.497\scriptsize$\pm$0.051 & 0.912\scriptsize$\pm$0.012 & \textbf{0.730\scriptsize$\pm$0.021} & 0.892\scriptsize$\pm$0.009 & \textbf{0.834\scriptsize$\pm$0.018} \\
 & JTT  & 0.940\scriptsize$\pm$0.020 & 0.481\scriptsize$\pm$0.065 & 0.923\scriptsize$\pm$0.012 & 0.719\scriptsize$\pm$0.027 & 0.875\scriptsize$\pm$0.016 & 0.789\scriptsize$\pm$0.036 \\
 & Ours & 0.829\scriptsize$\pm$0.041 & \textbf{0.840\scriptsize$\pm$0.060} & 0.875\scriptsize$\pm$0.035 & \underline{0.730\scriptsize$\pm$0.099} & 0.869\scriptsize$\pm$0.041 & \underline{0.805\scriptsize$\pm$0.032}\\
 \bottomrule
\end{tabular}

\label{tab:isic_tpr_abs}
\end{table*}

\section{Teacher sensitivity to shortcut features}
\begin{figure}[t]
    \centering
    \includegraphics[width=0.5\textwidth]{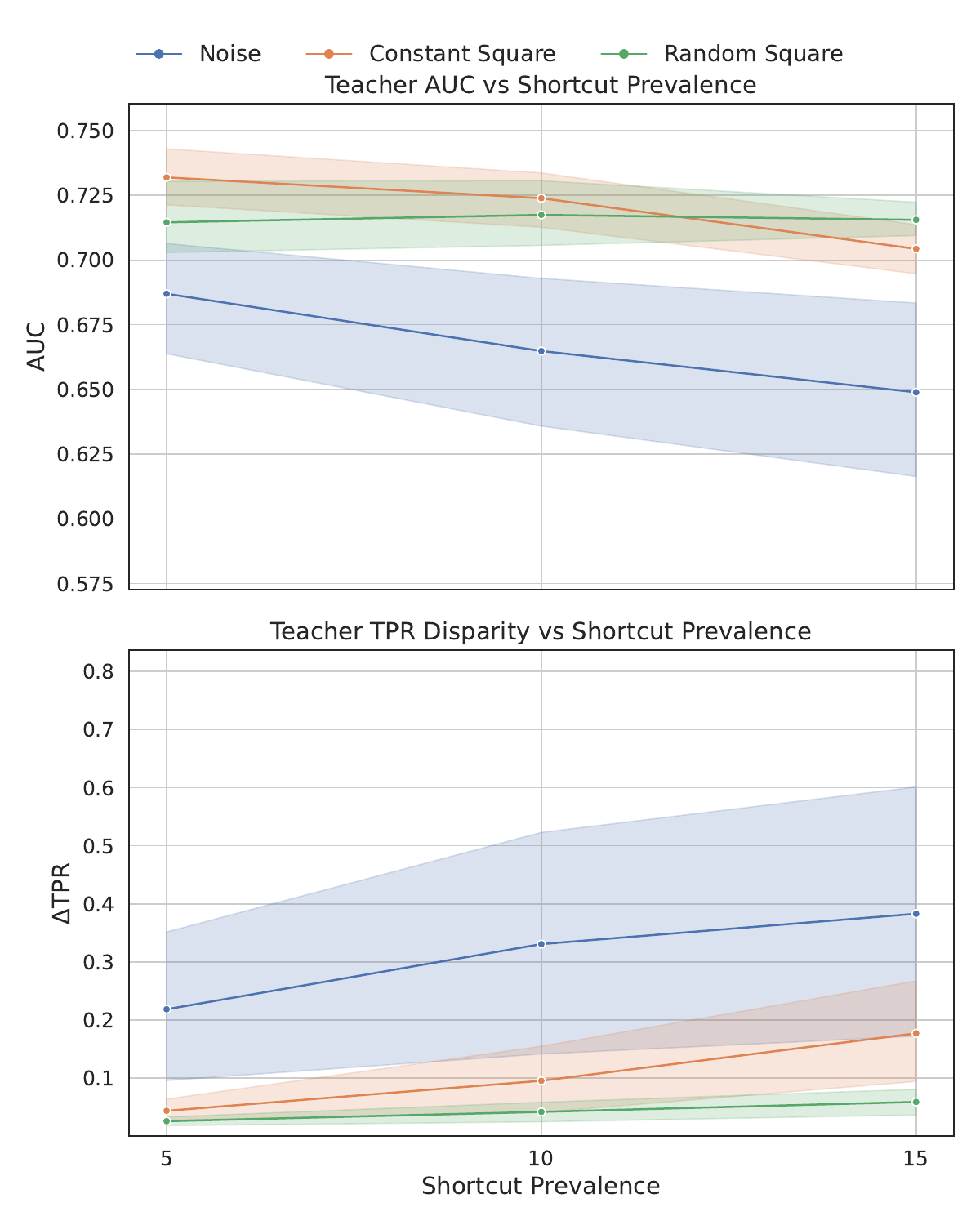}
    \caption{
    AUC and $\Delta$TPR of ResNet-18 teacher networks trained on 20\% subsets of the CheXpert, corrupted with shortcut features at various prevalence. 
    All test sets feature the same shortcut feature as is present in the train split, evenly distributed between samples belonging to each class, and therefore is no longer a useful predictive feature.
    }\label{fig:teacher_shortcut_corrupted}
\end{figure}

Here, we highlight the sensitivity of our teacher networks to corruption from shortcut features. Figure \ref{fig:teacher_shortcut_corrupted} supplements our findings in Section \ref{sec:shortcut_teacher} and supports our argument that practitioners should prioritize removing the most available (easily identifiable) shortcuts from the teacher training data as these cause the most significant harm.

\end{document}